\documentclass[10pt,journal,final,twocolumn]{IEEEtran}
\pdfoutput=1 
\usepackage{graphicx,times,amsmath,amssymb,cite,cases,mathrsfs,booktabs,multirow,subfigure,url}
\usepackage{algorithm,algorithmic,amsthm,ulem,latexsym,bm}
\usepackage[colorlinks,linkcolor=red,anchorcolor=blue,citecolor=red]{hyperref}

\begin{document}
%
\title{Domain Adaptation and Image Classification via Deep Conditional Adaptation Network}

\author{Pengfei~Ge, Chuan-Xian~Ren, Dao-Qing~Dai, Hong~Yan~\IEEEmembership{Fellow,~IEEE}
\thanks{P.F. Ge is with the School of Mathematics and Systems Science, Guangdong Polytechnic Normal University, Guangzhou, China.}
\thanks{C.X. Ren, and D.Q. Dai are with the Intelligent Data Center, School of Mathematics, Sun Yat-Sen University, Guangzhou, China.}
\thanks{H. Yan is with the Department of Electrical Engineering, City University of Hong Kong, Hong Kong.}}

\date{}
\IEEEcompsoctitleabstractindextext{%
\begin{abstract}%
Unsupervised domain adaptation (UDA) aims to generalize the supervised model trained on a source domain to an unlabeled target domain. Feature space alignment-based methods are widely used to reduce the domain discrepancy between the source and target domains. However, existing feature space alignment-based methods often ignore the conditional dependence between features and labels when aligning feature spaces, which may suffer from negative transfer. In this paper, we propose a novel UDA method, Deep Conditional Adaptation Network (DCAN), based on conditional distribution alignment of feature spaces. To be specific, we reduce the domain discrepancy by minimizing the Conditional Maximum Mean Discrepancy between the conditional distributions of deep features on the source and target domains, and extract the discriminant information from target domain by maximizing the mutual information between samples and the prediction labels. In addition, DCAN can be used to address a special scenario, Partial UDA, where the target domain category is a subset of the source domain category. Experiments on both UDA and Partial UDA show that DCAN achieves superior classification performance over state-of-the-art methods.
\end{abstract}

\begin{IEEEkeywords}
Domain adaptation, Image classification, Maximum mean discrepancy, Conditional distribution discrepancy, Mutual information.
\end{IEEEkeywords}}
\maketitle \IEEEdisplaynotcompsoctitleabstractindextext \IEEEpeerreviewmaketitle

\section{Introduction}\label{sect:introduction}

\IEEEPARstart{D}{eep} Convolutional Neural Networks (CNNs) have achieved great success in a variety of pattern recognition and computer vision applications~\cite{Krizhevsky2012ImageNet,chan2015pcanet}. A number of recent results show that when the deep networks are trained in large-scale datasets, the features show good generalization performance over a wide range of datasets and computer vision tasks~\cite{Donahue2014DeCAF,Yosinski2014How}. However, due to the dataset bias problem, test errors of these deep classification networks are large when the training set and test set have a significant gap in data distributions. Fine-tuning provides a straightforward way to reduce feature bias on deep networks~\cite{Oquab2014Learning}. Unfortunately, fine-tuning deep network parameters in a new dataset requires a significant amount of labeled data, which are not available in many scenarios. Therefore, it is necessary and important to design algorithms that can transfer discrimination features from a labeled source domain to another related but unlabeled domain.

\begin{figure}[htbp]
\centering{{\includegraphics[width=3.4in]{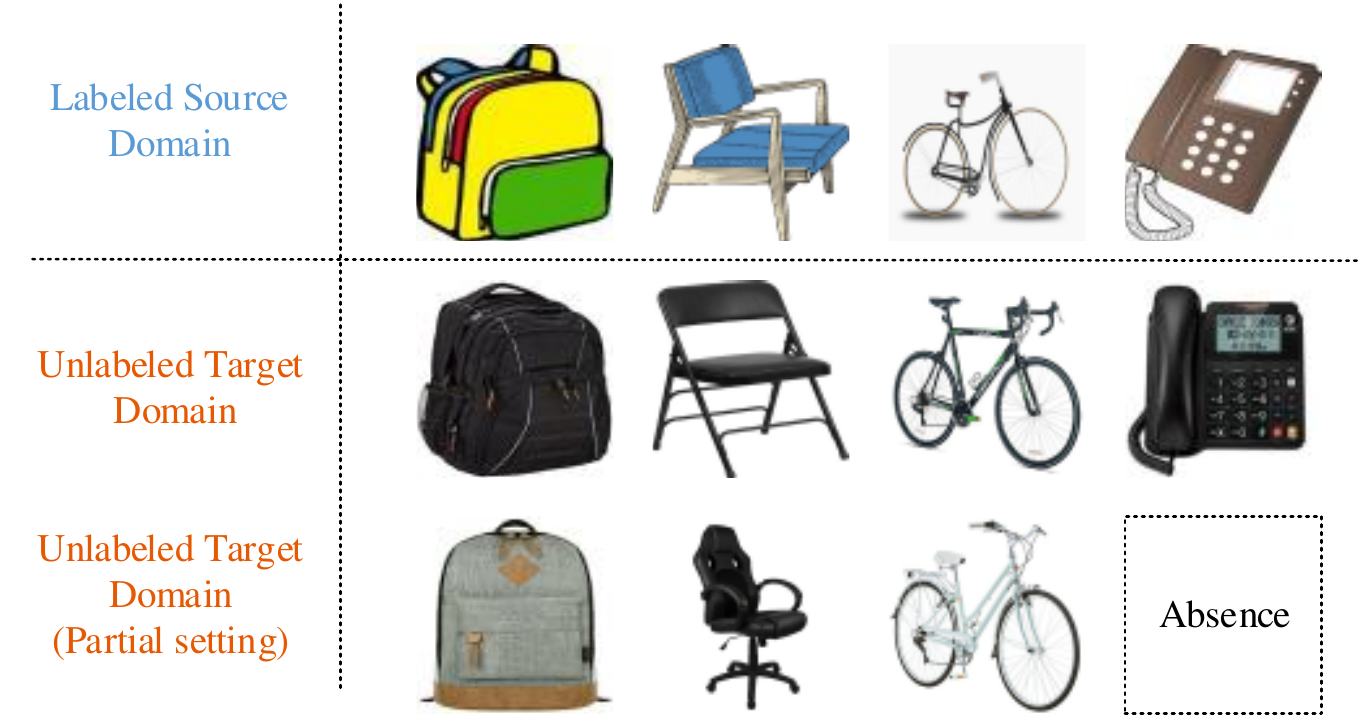}}
\caption{The diagrams of UDA and Partial UDA. In the UDA task, the source and target domains have different data distributions but the same category space. In Partial UDA task, the target domain category is a subset of the source domain category.}
\label{fig:1}} 
\end{figure}

To address this problem, a more practical task called unsupervised domain adaptation (UDA) has been studied recently. UDA generally contains two different but related datasets, i.e., a labeled source domain and an unlabeled target domain. The source and target domains have the same category space and learning tasks. The principal task of UDA is to transfer discriminative domain knowledge to solve the learning task in target domain~\cite{Pan2010A,Zhang2013Domain}. In particular, since the label information of target domain is agnostic, it is sometimes difficult to guarantee that the source and target domains share the same label space. In this paper, we also consider a special UDA scenario, namely Partial UDA~\cite{cao2018partialtransfer,cao2018partial}, where the target domain category is a subset of the source domain category. The diagrams of UDA and Partial UDA are shown in Figure~\ref{fig:1}.

Previous UDA methods are mainly based on shallow models~\cite{Zhang2013Domain,Donahue2013Semi,Ren2014Transfer,Ren2018Generalized}, which are roughly divided into two categories, i.e., instance-based methods and feature-based methods. Instance-based adaptation methods~\cite{Zhang2013Domain,Donahue2013Semi} reweigh samples in the source domain to better represent the target domain distribution, while feature-based methods~\cite{Ren2014Transfer,Ren2018Generalized} attempt to learn a shared and invariant feature space. However, limited by the model's representation capacity, the performance of these methods does not exceed the deep UDA approach.

In recent years, with the development of deep neural networks, more deep network-based models have been proposed to deal with UDA tasks~\cite{tzeng2014deep,long2015learning,ganin2016domain,tzeng2017adversarial}. These methods follow the idea that domain discrepancy becomes smaller in the deep feature space~\cite{Donahue2014DeCAF,Yosinski2014How}, thus domain adaptation can be better accomplished by matching the deep features of the source and target domains. Let $Z$ and $Y$ represent deep features and labels, respectively. To project data from different domains to a shared deep feature space, existing methods often rely on marginal distribution alignment, which reduces the discrepancy between $P^{s}(Z)$ and $P^{t}(Z)$. A common strategy is to minimize the maximum mean discrepancy (MMD)~\cite{tzeng2014deep,long2015learning} or introduce the adversarial training~\cite{ganin2016domain,tzeng2017adversarial}. However, the marginal distribution alignment-based method ignores the conditional dependence between features and labels when aligning the feature space, which may suffer from negative transfer. In particular, in the PDA task, aligning $P^s(Z)$ and $P^t(Z)$ will cause the target domain sample is assigned to a category that does not exist in the target domain~\cite{cao2018partial}. Therefore, the marginal distribution alignment-based methods cannot deal with (Partial) UDA problems well.

An effective strategy to address this problem is learn a conditional domain-invariant feature space by aligning the conditional distributions $P^{s}(Z|Y)$ and $P^{t}(Z|Y)$. Based on this motivation, some recent UDA methods reduce the difference between $P^{s}(Z|Y)$ and $P^{t}(Z|Y)$ by class-specific MMD~\cite{long2013transfer,wang2017balanced} or conditional adversarial generation network~\cite{long2018conditional}. However, class-specific MMD requires a large number of samples to estimate the MMD in each category, which makes it difficult to apply to deep models. Conditional adversarial generation network may suffer mode collapse and training instability. In addition, the conditional distribution alignment needs to use the label information of target domain, thus it is necessary to extract the discrimination information in target domain to improve the accuracy of the pseudo-labels. Some methods introduce an extra entropy regularization to extract discriminant information in the target domain~\cite{long2015learning,xu2018unsupervised}. However, the entropy regularization ignores the overall discriminant information, instead, only considers the discriminant information of a single sample. This manner may cause model degradation. How to effectively measure the difference between two conditional distributions and extract the discrimination information of unlabeled data in the deep model is the key to align the conditional distributions.

In this paper, we propose a new conditional distribution alignment-based domain adaptation method, named Deep Conditional Adaptation Network (DCAN), which can align effectively the conditional distributions by Conditional Maximum Mean Discrepancy (CMMD) and extract discriminant information from the source and target domains. CMMD can directly and efficiently measure the distance between two conditional distributions. Therefore, we use CMMD to measure the conditional distributions discrepancy between the source and target domains, and then minimize the CMMD loss to learn a conditional domain-invariant feature space, where the classifier trained on the source domain can correctly classify samples in the target domain. Compared with class-specific MMD, CMMD can be effectively estimated with fewer samples, thus it can be trained in a mini-batch manner and applied to the deep model. In addition, due to the absence of real target labels, we need to use the discriminant information in target domain to estimate CMMD. To extract the discriminant information in target domain, we introduce mutual information to measure how much information is represented by the predicted category. Mutual information can be expressed as the difference between the entropy and conditional entropy of the predicted category variable. By maximizing mutual information, we consider not only individual discrimination information, but also overall discrimination information. DCAN aims to learn a conditional domain-invariant feature space, thus it can also be extended to deal with Partial UDA. We evaluate DCAN on both UDA and Prtial UDA tasks, and the experiment results show that DCAN achieves state-of-the-art performance in all tasks.

Our contributions are summarized as follows.
\begin{enumerate}
  \item We propose a new deep domain adaptation approach, DCAN, to address both UDA and Partial UDA problems by aligning the conditional distributions of deep features and simultaneously extracting discriminant information from the source and target domains.
  \item We introduce CMMD to measure effciency the conditional distribution discrepancy between domains. CMMD can be estimated in a mini-batch of samples, thus DCAN can be trained in an end-to-end manner. To improve the accuracy of pseudo-labels, we introduce mutual information to extract the discriminant information on target domain.
  \item Extensive experiment results show that DCAN outperforms state-of-the-art UDA and Partial UDA methods on several image benchmarks.
\end{enumerate}

The rest of this paper is organized as follows. Section~\ref{sect:related-works} briefly reviews closely related works. In Section~\ref{sect:method}, we introduce DCAN in detail. Two algorithms dealing with both UDA and Partial UDA problems are presented. In Section~\ref{sect:experiments}, experiment results and analysis are presented, and DCAN is compared with several state-of-the-art methods. Section~\ref{sect:conclusion} concludes this paper.

\section{Related works}\label{sect:related-works}

In this section, we briefly review the visual domain adaption techniques from two directions, i.e., UDA and Partial UDA. Some preliminary of CMMD is also presented.

\subsection{Unsupervised Domain Adaptation}

Unsupervised Domain Adaptation is widely used in many areas of image processing, such as image classification~\cite{tzeng2014deep,long2015learning,ganin2016domain,long2018conditional}, image segmentation~\cite{zhang2019category,saito2018maximum,HUANG2022108384} and image binarization~\cite{CASTELLANOS2021108099}. In this subsection, we mainly review the application of UDA in image classification. MMD is the most common statistic used to measure domain discrepancy. Deep Domain Confusion (DDC)~\cite{tzeng2014deep} combines the classification loss with a MMD-based domain confusion loss to learn a domain-invariant feature space. Deep Adaptation Networks (DAN)~\cite{long2015learning} use MMD of all task-specific layers to reduce domain discrepancy, and it achieves more robust results than DDC. Residual transfer network (RTN)~\cite{long2016unsupervised} introduces additional residual structures and entropy minimization to implement classifier adaptation. These methods need to align the marginal distributions and ignore the conditional dependence between features and labels, which may cause negative transfer.

Recently, more conditional distribution alignment-based UDA methods are proposed. Joint distribution adaptation (JDA)~\cite{long2013transfer} introduces the class-specific MMD to estimate the domain discrepancy in each category. Balanced Distribution Adaptation (BDA)~\cite{wang2017balanced} extends this idea by weighting the importance of both marginal and conditional distribution adaptations. Manifold Embedded Distribution Alignment (MEDA)~\cite{wang2018visual} further introduces the dynamic distribution alignment to dynamically adjust the importance of both marginal and conditional distribution adaptations. Weighted Domain Adaptation Network (WDAN)~\cite{yan2017mind} proposes a weighted MMD to align feature distributions. It reweighs the samples in source domain to eliminate class weight bias across domains. Deep Subdomain Adaptation Network (DSAN)~\cite{DSAN} introduces a local maximum mean discrepancy to align the relevant subdomain distributions of different domains. There are several major differences between our DCAN and these methods. 1) Compared with WDAN, DCAN also uses the discriminant information contained in the target domain when aligning feature distributions, but does not need to calculate the weight. 2) All these methods use the sum of MMD on each class to estimate the conditional distribution divergence, while DCAN introduces CMMD to directly estimate the conditional distribution divergence. 3) DCAN introduces mutual information to extract the discriminant information in target domain, which effectively improves the accuracy of conditional distribution divergence estimation.

GAN is another common method for measuring domain discrepancy. Domain adversarial neural network (DANN)~\cite{ganin2016domain} introduces a domain discriminator to classify features of the source and target domains, while the feature extraction network learns domain-invariant features as much as possible to confuse the domain discriminator. Multi-adversarial domain adaptation (MADA)~\cite{pei2018multi} uses multiple domain discriminators to achieve fine-grained alignment of different data distributions. Conditional Domain Adversarial Network (CDAN)~\cite{long2018conditional} trains a conditional generative adversarial network by the discrimination information conveyed in the classifier. Generative Attention Adversarial Classification Network (GAACN)~\cite{GAACN} introduces an attention module in the process of adversarial learning to further minimize domain shifts.

\subsection{Partial Unsupervised Domain Adaptation}

Partial domain adaptation problem is originally proposed by~\cite{cao2018partialtransfer,cao2018partial}, where the target domain label space is a subset of the source domain label space. Recently, several methods are proposed to deal with partial domain adaptation problem. Importance Weighted Adversarial Net (IWAN)~\cite{zhang2018importance} introduces an additional domain classifier to calculate the weight of each sample in the source domain, and then uses GAN to align target domain features with this weighted source domain features. Partial Adversarial Domain Adaptation (PADA)~\cite{cao2018partial} uses the predicted distribution on the target domain to weight the class space of the source domain and then aligns feature space via GAN in the shared label space. Hard Adaptive Feature Norm (HAFN)~\cite{xu2018unsupervised} requires that the feature norms of the source and target domains be close to an identical larger value to complete domain adaptation, thus the assumption of identical label space can be ignored. Example Transfer Network (ETN)~\cite{cao2019learning} introduces a transferability quantifier to quantify the transferability of source examples, and reduces the negative influence of source outlier examples. Reinforced transfer network (RTNet)~\cite{chen2020selective} extends this idea by applying reinforcement learning to learn a reinforced data selector to select source outlier classes. Jigsaw~\cite{Jigsaw} address both UDA and Partial UDA problems by introducing self-supervised learning to learn the semantic labels.

\subsection{Conditional Maximum Mean Discrepancy}

In this subsection, we briefly review the basic principles of CMMD. Let $\phi$ and $\psi$ denote the nonlinear mappings for $Y$ and $Z$, respectively. For the conditional distribution $P(Z|Y)$, given a sample $\textbf{y}$, the conditional kernel mean embedding $\mu_{Z|\textbf{y}}$ can be defined as~\cite{Song2009Hilbert},
\begin{equation}\label{eq:cond-distr}
\mu_{Z|\textbf{y}}=\mathbb{E}_{Z|\textbf{y}}[\psi(Z)]=\int_\Omega\psi(z)dP(Z|\textbf{y})=C_{Z|Y}\phi(\textbf{y}),
\end{equation}
where $C_{Z|Y}$ denotes the conditional embedding of $P(Z|Y)$, which can be calculated as,
\begin{equation}\label{eq:cond-embed}
C_{Z|Y}=C_{YZ}C_{YY}^{-1}.
\end{equation}
$C_{YZ}$ denotes a cross-covariance operator~\cite{Baker1973Joint}, i.e.,
\begin{equation}\label{eq:cross-covariance}
  C_{YZ}=\mathbb{E}_{YZ}[\phi (Y) \otimes\psi (Z)]-\mu_Y\otimes\mu_Z,
\end{equation}
where $\otimes$ is the tensor product operator. Given a dataset $\mathcal{D}_{ZY}={(\textbf{z}_{i},\textbf{y}_{i})}_{i=1}^{n}$ from $P(Z|Y)$, $C_{Z|Y}$ can be estimated by $\widehat{C}_{Z|Y}=\Psi(\mathcal{L}+\lambda\textbf{I}) ^{-1}\Phi^\top$, where $\Psi= [\psi (\textbf{z}_1),\cdots, \psi (\textbf{z}_n)] $, $\Phi= [\phi(\textbf{y}_1),\cdots, \phi(\textbf{y}_n)] $, $\mathcal{L}=\Phi^\top\Phi$ is the kernel function, and $\lambda$ is a positive regularization parameter~\cite{ren2016conditional}.

Based on Eqs.~\eqref{eq:cond-distr} and \eqref{eq:cond-embed}, the conditional distributions are projected to a series of points in the RKHS. Since $\phi(\textbf{y})$ is a constant vector for a fixed $\textbf{y}$, the distance between the conditional kernel mean embeddings can be calculated by comparing the difference between the conditional embeddings. As a result, the CMMD between two conditional distribution $P(Z_{1}|Y)$ and $P(Z_{2}|Y)$ is defined as $\| C_{Z_{1}|Y}-C_{Z_{2}|Y}\| _{\mathcal{F\otimes G}}^2$.

\begin{figure*}[htbp]
\centering{{\includegraphics[width=6.4in]{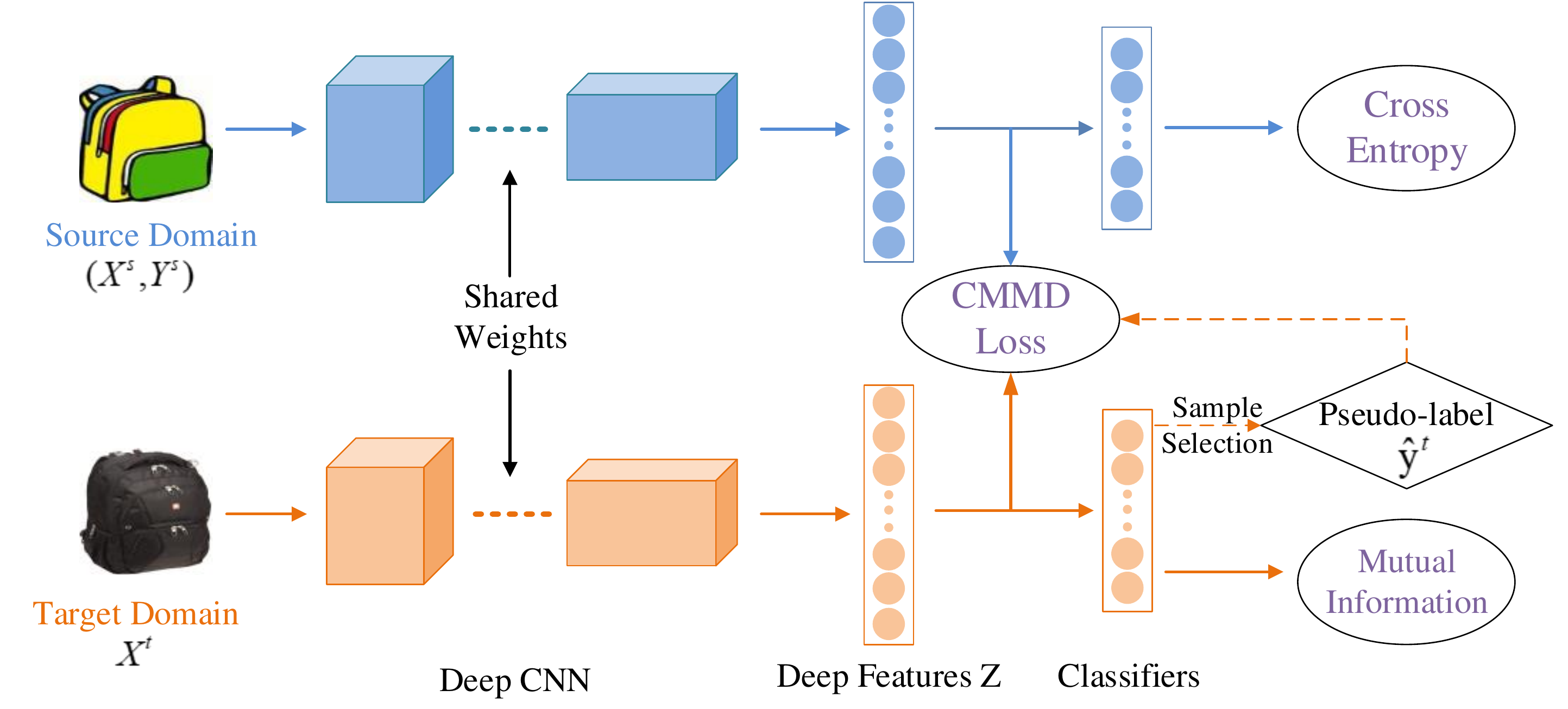}}
\caption{An overview of DCAN. The source and target domains share network weights of the Deep CNN and Classifier. We obtain the deep features by the Deep CNN, and then input these deep features into the Classifier to get predicted probabilities, which are used to calculate the cross-entropy loss and mutual information. At the same time, we select high-confidential samples in the target domain to assign pseudo-labels. The deep features with the labels and pseudo-labels are then used to estimate the CMMD loss to achieve conditional distributions alignment.}
\label{fig:flowchart}} 
\end{figure*}

\section{Method}\label{sect:method}

We first introduce the motivation and notations in Section~\ref{sect:MN}. Then, two important modules in DCAN, i.e., the conditional distribution matching module and the discriminant information extraction module, are introduced in Sections~\ref{sect:CDA} and~\ref{sect:DIE}, respectively. The final objective function and optimization algorithm of DCAN in Section~\ref{sect:DT}. In addition, Section~\ref{sect:PUDA} extends DCAN to address the Partial UDA problem.

\subsection{Motivation and Notations}\label{sect:MN}

The deep feature-level UDA methods assumes that there exists a deep feature space shared by the source domain and the target domain. In this feature space, both the labeled source samples and the unlabeled target samples can be classified as correctly as possible. Generally, this shared feature space should satisfy to the following two characteristics: 1)~The feature space should be domain-invariant for each category, which also means conditional distribution alignment. If the source and target domains have the same conditional distribution, then the classifier trained on the source domain can correctly classify samples in the target domain. 2)~The feature space should be able to extract discrimination information of the source domain and target domain simultaneously. This would be helpful to alleviate over-fitting on the source domain, thus learning a more generalized predictive model.

In order to learn a feature space which has the two properties as shown above, DCAN contains two interdependent modules: conditional distribution matching and discriminant information extraction. On the one hand, CMMD is used to measure the distance between conditional distributions of the source and target domains, and then the conditional distributions are aligned by minimizing CMMD. On the other hand, in addition to extracting discrimination information of the source domain using the cross-entropy loss, more representative features are captured from the target domain by maximizing the mutual information between samples and the prediction labels.


Several important notations are listed here. In the UDA task, there are generally two related but different datasets, i.e., the source domain $\mathcal{D}_{s}={(\textbf{x}_{i}^{s},\textbf{y}_{i}^{s})}_{i=1}^{n_{s}}$ with $n_{s}$ labeled samples and the target domain $\mathcal{D}_{t}={(\textbf{x}_{i}^{t})}_{i=1}^{n_{t}}$ with $n_{t}$ unlabeled samples, where $\textbf{y}_{i}$ denotes a one-hot encoding vector of the category. Let $P^{s}(X, Y)$ and $P^{t}(X, Y)$ represent the joint distribution on the source domain and target domain, respectively. Let $\mathcal{C}_{s}=\{1,\cdots,c_{s}\}$ and $\mathcal{C}_{t}=\{1,\cdots,c_{t}\}$ denote the label space of $\mathcal{D}_{s}$ and $\mathcal{D}_{t}$, respectively. In this paper, we first consider a common UDA scenario where $\mathcal{D}_{s}$ and $\mathcal{D}_{t}$ have the same label space, i.e., $\mathcal{C}_{s}=\mathcal{C}_{t}$. Furthermore, we also consider the Partial UDA task, which requires $\mathcal{C}_{t}\subset \mathcal{C}_{s}$.

\subsection{Conditional Distribution Alignment by CMMD}\label{sect:CDA}

In this section, we design the conditional distribution matching module with CMMD. The CMMD metric can project the conditional distribution into the Reproducing Kernel Hilbert Space (RKHS), and measure the distance between $P^{s}(Z|Y)$ and $P^{t}(Z|Y)$ by calculating the distance between their projections.

Let $\phi$ and $\psi$ denote the nonlinear mappings for $Y$ and $Z$, respectively. Then, the kernel mean embedding of $P(Y)$ and $P(Z)$ can be denoted by $\mu_Y=\mathbb{E}_Y[\phi(Y)]=\int_{\Omega}\phi(Y)dP(Y)$ and $\mu_Z=\mathbb{E}_Z[\psi(Z)]$, respectively. We can project a marginal distribution to a point in the RKHS via kernel mean embedding. Similarly, we can project the conditional distribution to a series of points in the RKHS via the conditional kernel mean embedding. Based on Eqs.~\eqref{eq:cond-distr}, given a sample $\textbf{y}$, the conditional kernel mean embedding of conditional distributions $P^{s}(Z|Y)$ and $P^{t}(Z|Y)$ can be defined as,
\begin{equation}\label{eq:cond-distr1}
\begin{aligned}
\mu^{s}_{Z|\textbf{y}}=\mathbb{E}_{Z^{s}|\textbf{y}}[\psi(Z)]=\int_\Omega\psi(z)dP^{s}(Z|\textbf{y})=C^{s}_{Z|Y}\phi(\textbf{y}),\\
\mu^{t}_{Z|\textbf{y}}=\mathbb{E}_{Z^{t}|\textbf{y}}[\psi(Z)]=\int_\Omega\psi(z)dP^{t}(Z|\textbf{y})=C^{t}_{Z|Y}\phi(\textbf{y}),
\end{aligned}
\end{equation}
where $C^{s}_{Z|Y}$ and $C^{t}_{Z|Y}$ denote the conditional embedding of $P^{s}(Z|Y)$ and $P^{t}(Z|Y)$, respectively.

To calculate the distance between $P^{s}(Z|Y)$ and $P^{t}(Z|Y)$ on the RKHS, we can calculate the distance between $\mu^{s}_{Z|\textbf{y}}$ and $\mu^{t}_{Z|\textbf{y}}$ by fixing $Y=\textbf{y}$. Based on the theory of CMMD, this distance can be calculated by comparing the difference between $C^{s}_{Z|Y}$ and $C^{t}_{Z|Y}$. Therefore, the CMMD loss between $P^{s}(Z|Y)$ and $P^{t}(Z|Y)$ as be written as,
\begin{equation}\label{eq:cmmd}
  L_{\rm{CMMD}}=\| C^{s}_{Z|Y}-C^{t}_{Z|Y}\| _{\mathcal{F\otimes G}}^2.
\end{equation}
By minimizing the CMMD loss, the difference between conditional distributions of $\mathcal{D}_{s}$ and $\mathcal{D}_{t}$ will be reduced. In particular, based on Theorem 3 of~\cite{ren2016conditional}, when $L_{\rm{CMMD}}$ reaches its minimum value, $P^{s}(Z|Y)=P^{t}(Z|Y)$ for each fixed $\textbf{y}$.

In practice, we need to estimate the CMMD loss in a batch-wise manner. To obtain the consistent estimator, we randomly sample two datasets $\mathcal{D}^{s}_{ZY}={(\textbf{z}_{i}^{s},\textbf{y}_{i}^{s})}_{i=1}^{n}$ and $\mathcal{D}^{t}_{ZY}={(\textbf{z}_{i}^{t},\textbf{y}_{i}^{t})}_{i=1}^{n}$ from $P^{s}(Z|Y)$ and $P^{t}(Z|Y)$, respectively. Then, an empirical estimation of CMMD between the conditional distribution of the source and target domains can be written as
\begin{equation}\label{eq:empir-cmmd}
  \begin{aligned}
	\widehat{L}_{\rm{CMMD}}=&\|\widehat{C}_{Z|Y}^s-\widehat{C}_{Z|Y}^t\|_{\mathcal{F\otimes G}}^2\\
	=&\|\Psi_s(\mathcal{L}_s+\lambda\textbf{I})^{-1}\Phi_s^\top-\Psi_t(\mathcal{L}_t+\lambda \textbf{I})^{-1}\Phi_t^\top\|_{\mathcal{F\otimes G}}^2\\ =&\mbox{Tr}(\mathcal{L}_s\widetilde{\mathcal{L}}_s^{-1}\mathcal{K}_s\widetilde{\mathcal{L}}_s^{-1})+\mbox{Tr}(\mathcal{L}_t\widetilde{\mathcal{L}}_t^{-1}\mathcal{K}_t\widetilde{\mathcal{L}}_t^{-1})\\
&-2\cdot \mbox{Tr}(\mathcal{L}_{ts}\widetilde{\mathcal{L}}_s^{-1}\mathcal{K}_{st}\widetilde{\mathcal{L}}_t^{-1}),
	\end{aligned}
\end{equation}
where $\Psi_s=[\psi (\textbf{z}_1^s),\cdots, \psi (\textbf{z}_n^s)]$, $\Phi_s=[\phi (\textbf{y}_1^s),\cdots, \phi (\textbf{y}_n^s)]$, $\mathcal{K}_s=\Psi_s^\top\Psi_s$, $\mathcal{L}_s=\Phi_s^\top\Phi_s$, $\widetilde{\mathcal{L}}_s=\mathcal{L}_s+\lambda \textbf{I}$. Accordingly, $\Psi_t$, $\Phi_t$, $\mathcal{K}_t$ $\mathcal{L}_t$ and $\widetilde{\mathcal{L}}_t$ are defined on dataset $\mathcal{D}_{ZY}^t$ in a similar way. $\mathcal{K}_{st}=\Psi_s^\top\Psi_t$, $\mathcal{L}_{ts}=\Phi_t^\top\Phi_s$.

Notice that the label information in the target domain is required for estimating the $L_{\rm{CMMD}}$, which cannot be satisfied in the UDA tasks. As with some UDA methods~\cite{sener2016learning,Hu2018duplex}, we obtain pseudo-labels of samples with high confidence in the target domain. In order to improve the representation capacity of the model and the accuracy of pseudo-labels, we use a deep CNN network as feature extractor and a fully-connected network with softmax activation as classifier. Figure~\ref{fig:flowchart} shows the network structure of DCAN. During the training process, we first input the samples of the source and target domains into the deep CNN to obtain the deep features, which are then input into the classifier to obtain the predicted labels. The predicted distribution $\hat{\textbf{y}}_{i}^{t}$ for any sample $\textbf{x}_{i}^{t}$ in the target domain is input into a sample selection program. If the maximum value of $\hat{\textbf{y}}_{i}^{t}$ is larger than a threshold $\gamma_{0}$, we consider this sample as high confidence and then assign this sample corresponding pseudo-label. Finally, the features and pseudo-labels of the selected target domain samples and the features and labels of the source domain samples are combined to calculate the CMMD loss. For those samples with low-confidence, they are not used to calculate the CMMD loss, but to extract the discriminant information in target domain, which will be introduced in detail in the next subsection.

\subsection{Discriminant Information Extraction}\label{sect:DIE}

In Section \ref{sect:CDA}, we proposed the CMMD-based conditional distribution matching module, which helps to achieve conditional distribution alignment in the shared feature space. On this basis, we expect to learn a more effective model by simultaneously extracting discriminant information in the source and target domains.

As shown in Figure \ref{fig:flowchart}, we connect a shared classifier behind the deep CNN to predict the class distribution. Let $P(\hat{Y}|X)$ denote the predicted class distribution. We use cross entropy loss as the classification loss function of the source domain. Suppose there is a dataset with batch-size $n$, i.e., $\mathcal{D}^{s}_{batch}=\{(\textbf{x}_{i}^{s},\textbf{y}_{i}^{s})\}_{i=1}^{n}$, sampled from the source domain $\mathcal{D}_{s}$. The cross-entropy loss $L_{SC}$ can be estimated as
\begin{equation}\label{eq:cross entropy}
	\widehat{L}_{SC}=\frac{1}{n}\sum_{i=1}^{n}\left [ -P(\textbf{y}_{i}^{s}|\textbf{x}_{i}^{s})\log P(\hat{\textbf{y}}_{i}^{s}|\textbf{x}_{i}^{s}) \right ].
\end{equation}

Now we focus on extracting discriminant information from the target domain, in which all the samples are unlabeled, thus it is difficult to extract classification information by a supervised method. In probability theory and information theory, the mutual information between two variables $X$ and $Y$ measures how much information $Y$ is contained in $X$. Therefore, we can extract more representative features in the target domain by maximizing the mutual information between $X_{t}$ and $\hat{Y}_{t}$. The mutual information loss $L_{MI}$ of the target domain is defined as the negative value of $I(X_{t},\hat{Y}_{t})$, i.e.,
\begin{equation}\label{eq:mutual information}
	L_{MI}=-I(X_{t},\hat{Y}_{t})=H(\hat{Y}_{t}|X_{t})-H(\hat{Y}_{t}),
\end{equation}
in which $H(\hat{Y}_{t})$ denotes the information entropy of $P^{t}(\hat{Y})$ and $H(\hat{Y}_{t}|X_{t})$ denotes the conditional entropy of $P^{t}(\hat{Y}|X)$. By minimizing the mutual information loss, we simultaneously minimize the entropy of category conditional distribution and maximize the entropy of category marginal distribution. In the UDA literature~\cite{long2015learning,xu2018unsupervised}, the entropy of conditional distribution, also known as low-density separation criterion, has been used to constrain the classification boundary trained on the source domain through the low density region of the target domain feature space to prevent over-fitting. However, minimizing the entropy of the conditional distribution may result in excessive aggregation of the samples, further result in multiple categories of samples being grouped into one category. Different from these methods, we also maximize the entropy of marginal distribution to alleviate class imbalance problem and extract more discrimination features.

In practice, given a dataset with batch-size $n$, i.e., $\mathcal{D}^{t}_{batch}=\{\textbf{x}_{i}^{t}\}_{i=1}^{n}$, sampled from the target domain $\mathcal{D}_{t}$, we can obtain the corresponding prediction classification probability $P(\hat{\textbf{y}}_{i}^{t}|\textbf{x}_{i}^{t})$ through the deep network. Then, the mutual information loss can be estimated as
\begin{equation}\label{eq:i_e} \widehat{L}_{MI}=\frac{1}{n}\sum_{i=1}^{n}H(\hat{\textbf{y}}_{i}^{t}|\textbf{x}_{i}^{t})-H[\frac{1}{n}\sum_{i=1}^{n}P(\hat{\textbf{y}}_{i}^{t}|\textbf{x}_{i}^{t})].
\end{equation}

\subsection{DCAN \& Training Procedure}\label{sect:DT}

Now we can define the objective function of DCAN by combining these three loss functions~\eqref{eq:cmmd},~\eqref{eq:cross entropy},~\eqref{eq:mutual information} as follow,
\begin{equation}\label{eq:dcan}
	\min_{\textbf{w}} L_{SC}+ \lambda _{0} L_{\rm{CMMD}} + \lambda _{1} L_{MI},
\end{equation}
where $\textbf{w}$ denotes all the network parameters in the deep network, $\lambda _{0}$ and $\lambda _{1}$ are hyper-parameters.

DCAN consists of two successive training phases. First, to ensure that the deep network can obtain more high-confidence pseudo-labels at the initial transfer, we pre-train the deep network and classifier using the samples in the source domain. Next, we minimize the objective function~\eqref{eq:dcan} by mini-batch stochastic gradient descent method. Based on the chain rule, the gradient of the loss function on a mini-batch can be written as
\begin{equation}
\begin{aligned}
\sum_{i=1}^{n}\Bigg[&(\frac{\partial \hat{\textbf{y}}_{i}^{s}}{\partial \textbf{w}})^\top \frac{\partial \widehat{L}_{SC}}{\partial \hat{\textbf{y}}_{i}^{s}}+ \lambda _{1}(\frac{\partial \hat{\textbf{y}}_{i}^{t}}{\partial \textbf{w}})^\top\frac{\partial \widehat{L}_{\rm{MI}}}{\partial \hat{\textbf{y}}_{i}^{t}}\\ &+ \lambda_{0}\sum_{j=1}^{n}\left(\frac{\partial \widehat{L}_{\rm{CMMD}}}{\partial z_{ij}^{s}} \frac{\partial z_{ij}^{s}}{\partial \textbf{w}} + \frac{\partial \widehat{L}_{\rm{CMMD}}}{\partial z_{ij}^{t}} \frac{\partial z_{ij}^{t}}{\partial \textbf{w}}\right) \Bigg],
\end{aligned}
\end{equation}
where $z_{ij}$ denotes the $j$-th dimensional feature representation of $\textbf{x}_{i}$. For simplicity, the gradients of $\widehat{L}_{SC}$ and $\widehat{L}_{MI}$ are omitted here. For CMMD, Eq.~\eqref{eq:empir-cmmd} can be expressed as
\begin{equation}
\widehat{L}_{\rm{CMMD}}=\mbox{Tr}(G_s\mathcal{K}_{s})+\mbox{Tr}(G_{t}\mathcal{K}_t)-2\cdot \mbox{Tr}(G_{ts}\mathcal{K}_{st}),
\end{equation}
where $G_s = \widetilde{\mathcal{L}}_s^{-1}\mathcal{L}_s\widetilde{\mathcal{L}}_s^{-1}$, $G_t = \widetilde{\mathcal{L}}_t^{-1}\mathcal{L}_t\widetilde{\mathcal{L}}_t^{-1}$ and $G_{ts} = \widetilde{\mathcal{L}}_t^{-1}\mathcal{L}_{ts}\widetilde{\mathcal{L}}_s^{-1}$. Since $G_s$, $G_t$ and $G_{ts}$ are constant matrices, $\frac{\partial \widehat{L}_{\rm{CMMD}}}{\partial z_{ij}^{s}}$ can be calculated as,
\begin{equation}
\begin{aligned}
\frac{\partial \widehat{L}_{\rm{CMMD}}}{\partial z_{ij}^{s}} &= \frac{\partial \mbox{Tr}(G_s\mathcal{K}_{s})}{\partial z_{ij}^{s}} - 2\cdot\frac{\partial \mbox{Tr}(G_{ts}\mathcal{K}_{st})}{\partial z_{ij}^{s}}\\
&=\mbox{Tr}(G_s\frac{\partial \mathcal{K}_{s}}{\partial z_{ij}^{s}}) - 2\cdot\mbox{Tr}(G_{ts}\frac{\partial \mathcal{K}_{st}}{\partial z_{ij}^{s}}).
\end{aligned}
\end{equation}
Similarly, $\frac{\partial \widehat{L}_{\rm{CMMD}}}{\partial z_{ij}^{t}}$ can also be computed. All the gradients can be easily computed in the Pytorch framework. The overall algorithm of DCAN is summarized in Algorithm \ref{Alg1}.

\begin{algorithm}[htbp]
		\caption{DCAN for UDA}
		\label{Alg1}
		\begin{algorithmic}[1]
            \REQUIRE{$\mathcal{D}_{s}=\{(\textbf{x}_{i}^{s},\textbf{y}_{i}^{s})\}_{i=1}^{n_{s}}$, $\mathcal{D}_{t}=\{\textbf{x}_{i}^{t}\}_{i=1}^{n_{t}}$}.
            \ENSURE{The network parameters $\textbf{w}$}.
			\STATE{Pre-train the deep network with samples in $\mathcal{D}_{s}$.}
			\WHILE{not converged}
			\STATE{Random sample a mini-batch $\mathcal{D}^{s}_{batch}$ and $\mathcal{D}^{t}_{batch}$ from $\mathcal{D}_s$ and $\mathcal{D}_t$, respectively;}
			\STATE{Generate $\hat{y}_{i}^{t}$ for each sample $\textbf{x}_{i}^{t}\in\mathcal{D}^{t}_{batch}$ by current classifier, and predict pseudo-labels for samples with high confidence;}
			\STATE{Estimate the CMMD loss, cross-entropy loss and mutual information loss by Eqs.~\eqref{eq:empir-cmmd},~\eqref{eq:cross entropy}, and~\eqref{eq:i_e}, respectively;}
            \STATE{Compute the gradients of Eqs.~\eqref{eq:empir-cmmd},~\eqref{eq:cross entropy}, and~\eqref{eq:i_e} w.r.t. $\textbf{w}$ on $\mathcal{D}^{s}_{batch}$ and $\mathcal{D}^{t}_{batch}$;}
			\STATE{Update $\textbf{w}$ by gradient descend to minimize Eq.~\eqref{eq:dcan};}
			\ENDWHILE
		\end{algorithmic}
\end{algorithm}

\subsection{Extension to Deal with the Partial UDA Problem}\label{sect:PUDA}

In this subsection, we extend the DCAN algorithm to accommodate the Partial UDA tasks. As shown above, the objective of CDAN contains three loss functions, i.e., CMMD loss for conditional distribution alignment, cross-entropy loss for extracting discriminant information of $\mathcal{D}_s$, and mutual information loss for extracting more representative information of $\mathcal{D}_t$. The CMMD loss and cross-entropy loss are unaffected by the change of the label space and thus can be used in the Partial UDA task. The mutual information loss can be decomposed into the difference between $H(\hat{Y}_{t}|X_{t})$ and $H(\hat{Y}_{t})$. In the partial setting, we cannot accurately estimate $P(Y_{t})$ by $P(\hat{Y}_{t})$, Thus the mutual information loss needs to be modified to accommodate the Partial UDA task.

\begin{figure*}[htb]
\centering{{\includegraphics[width=6.8in]{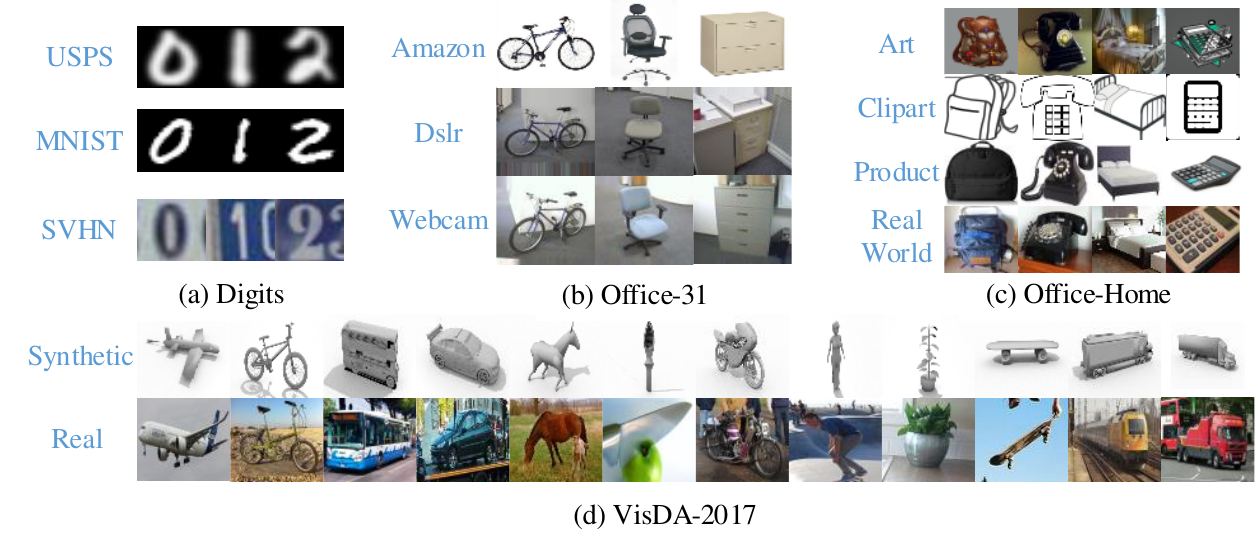}}
\caption{Sample images of four datasets, Digit Datasets, Office-31, Office-Home and VisDA-2017. Samples in the same row belong to the same domain, and samples in the same column belong to the same class.}\label{fig:sample}} 
\end{figure*}

In the partial setting, maximizing $H(P(\hat{Y}_{t}))$ will cause the target domain samples to be distributed evenly to each category of $\mathcal{D}_s$, which perhaps result in negative transfer. To solve this problem, one method is to remove the constraint on $H(P(\hat{Y}_{t}))$. However, as we mentioned above, minimizing $H(P(\hat{Y}_{t}|X_{t}))$ may lead to excessive aggregation of samples in $\mathcal{D}_t$, and maximizing $H(P(\hat{Y}_{t}))$ can help to mitigate this problem. To avoid the excessive aggregation of target domain samples, we introduce a threshold $\gamma_{1}$ to control the effect of $H(P(\hat{Y}_{t}))$. When $H(P(\hat{Y}_{t}))<\gamma_{1}$, we maximize $H(P(\hat{Y}_{t}))$ to ensure class balance, and when $H(P(\hat{Y}_{t}))>\gamma_{1}$, we remove the effect of $H(P(\hat{Y}_{t}))$. The mutual information loss under the partial setting is
\begin{equation}\label{eq:pmi}
	L_{MI}^{P}=H(\hat{Y}_{t}|X_{t})-\min\{H(\hat{Y}_{t}),\gamma_{1}\}.
\end{equation}
The objective function of Partial DCAN is then formulated by
\begin{equation}\label{eq:pdcan}
	\underset{\textbf{w}}\min\:\: L_{SC}+ \lambda _{0} L_{\rm{CMMD}} + \lambda _{1} L_{MI}^{p}.
\end{equation}
Eq.~\eqref{eq:pdcan} can be optimized in an end-to-end manner, and the main steps for addressing Partial UDA are shown in Algorithm \ref{Alg2}.

\begin{algorithm}[htbp]
		\caption{DCAN for Partial UDA}
		\label{Alg2}
		\begin{algorithmic}[1]
            \REQUIRE{$\mathcal{D}_{s}=\{(\textbf{x}_{i}^{s},\textbf{y}_{i}^{s})\}_{i=1}^{n_{s}}$, $\mathcal{D}_{t}=\{\textbf{x}_{i}^{t}\}_{i=1}^{n_{t}}$}.
            \ENSURE{The network parameters $\textbf{w}$}.
			\STATE{Pre-train the deep network with samples in $\mathcal{D}_{s}$.}
			\WHILE{not converged}
			\STATE{Random sample a mini-batch $\mathcal{D}^{s}_{batch}$ and $\mathcal{D}^{t}_{batch}$ from $\mathcal{D}_s$ and $\mathcal{D}_t$, respectively;}
			\STATE{Generate $\hat{y}_{i}^{t}$ for each sample $\textbf{x}_{i}^{t}\in\mathcal{D}^{t}_{batch}$ by current classifier, and predict pseudo-labels for samples with high confidence;}
			\STATE{Estimate the CMMD loss, cross-entropy loss and partial mutual information loss by Eqs.~\eqref{eq:empir-cmmd},~\eqref{eq:cross entropy}, and~\eqref{eq:pmi}, respectively;}
            \STATE{Compute the gradients of Eqs.~\eqref{eq:empir-cmmd},~\eqref{eq:cross entropy}, and~\eqref{eq:pmi} w.r.t. $\textbf{w}$ on $\mathcal{D}^{s}_{batch}$ and $\mathcal{D}^{t}_{batch}$;}
			\STATE{Update $\textbf{w}$ by gradient to minimize Eq.~\eqref{eq:pdcan};}
			\ENDWHILE
		\end{algorithmic}
\end{algorithm}

\section{Experiment results and analysis}\label{sect:experiments}

In this section, we evaluate DCAN method on four visual datasets. We first test its classification accuracy on UDA and Partial UDA tasks, and compare it with several state-of-the-art deep learning methods. Then, we evaluate its effectiveness from several views, i.e., parameter sensitivity, evaluation of each component, convergence performance, impact of the number of classes, analysis of pseudo-labels, feature visualization and time complexity.

\subsection{Datasets and Experimental Setup}

All experiments are conducted on four visual datasets, i.e., Digit, Office-31, Office-Home and VisDA-2017, which are widely used to test UDA algorithms. Some image samples of these datasets are shown in Figure \ref{fig:sample}.

{\bfseries Digit Datasets} contains 10 categories of digital images (0-9) from three domains, i.e., MNIST~\cite{lecun1998gradient}, USPS~\cite{Denker1989neural} and SVHN~\cite{Netzer2011reading}. MNIST includes 70,000 handwritten digital images of 10 classes, and each image has $28\times 28$ size. USPS is also handwritten digital images dataset, which consists of 9,298 gray images with size of $16\times 16$. SVHN consists of 73257 color digital images with size of $3\times 32\times 32$, which are all captured from house numbers. We conduct three widely-used UDA tasks (i.e., MNIST$\rightarrow$USPS, USPS$\rightarrow$MNIST and SVHN$\rightarrow$MNIST) to evaluate the DCAN method.

{\bfseries VisDA-2017}~\cite{peng2017visda} is a challenging large-scale dataset, which contains more than 280,000 images in 12 categories from two very distinct domains, i.e., Synthetic and Real. Synthetic includes 152,397 renderings of 3-D models. Real includes consists of 55,388 real object images. We conduct the UDA task from Synthetic to Real.

{\bfseries Office-31}~\cite{Saenko2010Adapting} is the most popular dataset for UDA task, which collects 4,110 images of office supplies in 31 classes from three distinct domains: Amazon~(A) consisting of online web images, DSLR~(D) consisting of digital SLR camera images, and Webcam~(W) consisting of web camera images. In order to conduct the Partial UDA experiment, we follow the literature~\cite{cao2018partial} to use the common sub-dataset from Office-31 and Office -10 as target domain. We evaluate DCAN method across all six UDA tasks and six Partial UDA tasks.

{\bfseries Office-Home}~\cite{Venkateswara2017Deep} is a more challenging dataset for UDA task, which contains 15588 images of common objects in 65 categories from four different areas: Art (Ar), Clipart (Cl), Product (Pr) and Real-World (Rw). In the Partial UDA task, we follow the setting in literature~\cite{cao2018partial} to use all the samples of a domain as the source domain, and the samples from the first 25 categories (in alphabetic order) in another domain as the target domain. We conduct all 12 UDA tasks and 12 Partial UDA tasks to evaluate DCAN method.

In the MNIST$\rightarrow$USPS and USPS$\rightarrow$MNIST tasks, we resize the samples in USPS to $28\times 28$ to match the MNIST image size. For the SVHN$\rightarrow$MNIST task, we resize the MNIST images to $32\times 32$ and then extend these images to three channels to match the image size in SVHN. In the rest of the tasks, we use a unified data processing protocol. Specifically, for an image, we use the following three operations in turn: 1) resize this image to $3\times224\times224$, 2) convert this image in the range [0, 255] to a tensor in the range [0.0, 1.0], 3) normalize this tensor with mean [0.485, 0.456, 0.406] and standard deviation [0.229, 0.224, 0.225].

Some implementation details are summarized as follows. In the transfer tasks of the digital datasets, we use the simple modified LeNet~\cite{lecun1998gradient} used in literatures~\cite{ganin2016domain,tzeng2017adversarial} as the baseline network structure and randomly initialize the network weights. In the other transfer tasks, we follow the setting in benchmark approaches~\cite{long2016unsupervised,long2018conditional} and use the Pytorch-provided ResNet~\cite{he2016deep} pre-trained model as the baseline network structure (ResNet-101 for VisDA-2017, whereas ResNet-50 for others). In all experiments, we fix the hyper-parameters as $\lambda _{0}=0.1$, $\lambda _{1}=0.2$, $\gamma_{0}=0.95$, $\gamma_{1}=1.5$ and the batch-size as $n=32$. To estimate the CMMD loss, we use a mixture kernel function obtained from an average of five Gaussian kernels, where the bandwidth is set to 0.1, 1, 10, 100, and 1000, respectively. We replace the fully connected (FC) layer of ResNet with a FC network to accommodate the new category number. For digital transfer tasks, we use ADAM optimization algorithm~\cite{Kingma2014Adam} with learning rate 2e-4. For the other three datasets, we set a learning rate of 2e-4 for the FC-layers, while the other layers are 2e-5. We follow the standard training and evaluation settings in UDA, i.e., using the labeled samples of $\mathcal{D}_s$ and the unlabeled samples of $\mathcal{D}_t$ for training, and then testing the classification accuracy in $\mathcal{D}_t$. In each experiment, we repeat DCAN three times with different random initialization, and report the mean accuracy and standard deviation following~\cite{xu2018unsupervised}.

\begin{table}[]
\centering
\renewcommand{\tabcolsep}{0.34pc} 
\renewcommand{\arraystretch}{1.0} 
\caption{Comparison with state-of-the-art methods on the digit datasets for unsupervised domain adaptation.}
\label{table1}
\begin{tabular}{cccc}
\hline
Method    & MNIST$\rightarrow$USPS & USPS$\rightarrow$MNIST  &  SVHN$\rightarrow$MNIST\\
\hline
LeNet~\cite{lecun1998gradient} & 75.2$\pm$0.2 & 57.1$\pm$0.2 & 60.1$\pm$0.1\\
BDA~\cite{wang2017balanced}      & 88.56 & 82.8 &- \\
DANN~\cite{ganin2016domain}      	& 77.1$\pm$1.8 & 73.0$\pm$2.0& 73.9$\pm$0.1\\
DRCN~\cite{Ghifary2016Deep}       & 91.8$\pm$0.1 & 73.7$\pm$0.1 & 82.0$\pm$0.2\\
CoGAN~\cite{Liu2016Coupled}       & 91.2$\pm$0.8 & 89.1$\pm$0.8&- \\
WDAN~\cite{yan2017mind}        & 90.2$\pm$0.1 & 89.2$\pm$0.5& 80.5$\pm$2.5\\
GTA~\cite{Sankaranarayanan2017Generate}      & 92.8$\pm$0.9 & 90.8$\pm$1.3  & 92.4$\pm$0.9\\
MCD~\cite{saito2018maximum} & 92.1$\pm$0.8 & 94.1$\pm$0.3 & 94.2$\pm$2.6\\
GAACN~\cite{GAACN} &95.4 &98.3 &94.6\\
GDN~\cite{YANG2021107638}   & 94.1$\pm$0.1 &98.4$\pm$0.1 &85.0$\pm$1.0\\
DSAN~\cite{DSAN}  & 96.9$\pm$0.2 & 95.3$\pm$0.1  & 90.1$\pm$0.4\\
DCAN(ours) & {\bfseries 97.5}$\pm$0.1 & {\bfseries 98.5}$\pm$0.1 & {\bfseries 98.7}$\pm$0.1\\
\hline
\end{tabular}
\end{table}

\begin{table*}[htb]
\centering
\renewcommand{\tabcolsep}{0.9pc} 
\renewcommand{\arraystretch}{1} 
\caption{Comparison with state-of-the-art methods on Office-31 for unsupervised domain adaptation (ResNet-50).}
\label{table2}
\begin{tabular}{cccccccc}
\hline
Method      &  A$\rightarrow$W & D$\rightarrow$W & W$\rightarrow$D & A$\rightarrow$D & D$\rightarrow$A & W$\rightarrow$A & Avg\\
\hline
Resnet-50~\cite{he2016deep}   & 68.4$\pm$0.2 & 96.7$\pm$0.1 & 99.3$\pm$0.1 & 68.9$\pm$0.2 & 62.5$\pm$0.3 & 60.7$\pm$0.3 & 76.1\\
DAN~\cite{long2015learning}     	& 83.8$\pm$0.4 & 96.8$\pm$0.2 & 99.5$\pm$0.1 & 78.4$\pm$0.2 & 66.7$\pm$0.3 & 62.7$\pm$0.2 & 81.3\\
BDA~\cite{wang2017balanced}         & 83.4 & 97.5 & 99.6 & 80.5 & 67.7 & 68.7 & 82.9\\
DANN~\cite{ganin2016domain}        & 82.0$\pm$0.4 & 96.9$\pm$0.2 & 99.1$\pm$0.1 & 79.7$\pm$0.4 & 68.2$\pm$0.4 & 67.4$\pm$0.5 & 82.2\\
JAN~\cite{long2017deep}       	& 85.4$\pm$0.3 & 97.4$\pm$0.2 & 99.8$\pm$0.2 & 84.7$\pm$0.3 & 68.6$\pm$0.3 & 70.0$\pm$0.4 & 84.3\\
WDAN~\cite{yan2017mind}       	& 84.3$\pm$0.4 & 98.0$\pm$0.3 & 99.9$\pm$0.6 & 80.4$\pm$0.6 & 66.0$\pm$1.8 & 62.3$\pm$1.0 & 82.0\\
MADA~\cite{pei2018multi}        & 90.0$\pm$0.1 & 97.4$\pm$0.1 & 99.6$\pm$0.1 & 87.8$\pm$0.2 & 70.3$\pm$0.3 & 66.4$\pm$0.3 & 85.2\\
GTA~\cite{Sankaranarayanan2017Generate}       	& 89.5$\pm$0.5 & 97.9$\pm$0.3 & 99.8$\pm$0.4 & 87.7$\pm$0.5 & 72.8$\pm$0.3 & 71.4$\pm$0.4 & 86.6\\
CDAN+E~\cite{long2018conditional}      & {\bfseries 94.1}$\pm$0.1 & 98.6$\pm$0.1 & {\bfseries 100}$\pm$0.0 & 92.9$\pm$0.2 & 71.0$\pm$0.3 & 69.3$\pm$0.3 & 87.7\\
SAFN~\cite{xu2018unsupervised} & 88.8$\pm$0.4 & 98.4$\pm$0.0 & 99.8$\pm$0.0 & 87.7$\pm$1.3 & 69.8$\pm$0.4 & 69.7$\pm$0.2 & 85.7\\
GAACN~\cite{GAACN} &90.2 &98.4 &{\bfseries 100.0} &90.4 &67.4 &67.7 &85.6\\
DSAN~\cite{DSAN}         & 93.6$\pm$0.2 & 98.3$\pm$0.1 & {\bfseries 100}$\pm$0.0 & 90.2$\pm$0.7 & 73.5$\pm$0.5 & {\bfseries 74.8}$\pm$0.4 & 88.4\\
DCAN(ours)  & 93.2$\pm$0.3 & {\bfseries 98.7}$\pm$0.1 & {\bfseries 100}$\pm$0.0 & 91.6$\pm$0.4 & {\bfseries 74.6}$\pm$0.2 & 74.2$\pm$0.2 & {\bfseries 88.7}\\
\hline
\end{tabular}
\end{table*}

\begin{table*}[htb]
\centering
\renewcommand{\tabcolsep}{0.4pc} 
\renewcommand{\arraystretch}{1.0} 
\caption{Comparison with state-of-the-art methods on Office-Home for unsupervised domain adaptation (ResNet-50).}
\label{table4}
\begin{tabular}{cccccccccccccc}
\hline
Method      &  A$\rightarrow$C & A$\rightarrow$P & A$\rightarrow$R & C$\rightarrow$A & C$\rightarrow$P & C$\rightarrow$R & P$\rightarrow$A & P$\rightarrow$C & P$\rightarrow$R & R$\rightarrow$A & R$\rightarrow$C & R$\rightarrow$P & Avg\\
\hline
Resnet-50~\cite{he2016deep}   & 34.9 & 50.0 & 58.0 & 37.4 & 41.9 & 46.2 & 38.5 & 31.2 & 60.4 & 53.9 & 41.2 & 59.9 & 46.1\\
DAN~\cite{long2015learning}      	& 43.9 & 57.0 & 67.9 & 45.8 & 56.5 & 60.4 & 44.0 & 43.6 & 67.7 & 63.1 & 51.5 & 74.3 & 56.3 \\
BDA~\cite{wang2017balanced}       	& 46.4 & 64.8 & 67.8 & 42.1 & 60.9 & 61.7 & 49.7 & 43.3 & 71.2 & 59.0 & 50.5 & 77.1 & 57.9 \\
DANN~\cite{ganin2016domain}       	& 45.6 & 59.3 & 70.1 & 47.0 & 58.5 & 60.9 & 46.1 & 43.7 & 68.5 & 63.2 & 51.8 & 76.8 & 57.6 \\
JAN~\cite{long2017deep}       	& 45.9 & 61.2 & 68.9 & 50.4 & 59.7 & 61.0 & 45.8 & 43.4 & 70.3 & 63.9 & 52.4 & 76.8 & 58.3\\
WDAN~\cite{yan2017mind}       	& 47.1 & 69.1 & 74.0 & 58.0 & 64.9 & 68.8 & 53.7 & 45.6 & 75.6 & 67.4 & 53.3 & 80.4 & 63.2\\
CDAN+E~\cite{long2018conditional}      & 50.7 & 70.6 & 76.0 & 57.6 & 70.0 & 70.0 & 57.4 & 50.9 & 77.3 & 70.9 & 56.7 & 81.6 & 65.8\\
SAFN~\cite{xu2018unsupervised} & 52.0 & 71.7 & 76.3 & 64.2 & 69.9 & 71.9 & 63.7 & 51.4 & 77.1 & 70.9 & 57.1 & 81.5 & 67.3\\
GAACN~\cite{GAACN} &53.1 &71.5 &74.6 &59.9 &64.6 &67.0 &59.2 &53.8 &75.1 &70.1 &59.3 &80.9 &65.8\\
Jigsaw~\cite{Jigsaw} &47.7 &58.8 &67.9 &57.2 &64.3 &66.1 &56.2 &50.8 &75.1 &67.9 &55.6 &78.4 &62.2\\
GDN~\cite{YANG2021107638} &51.7 &71.5 &75.7 &51.9 &66.1 &68.6 &53.9 &49.0 &74.0 &64.0 &53.2 &78.6 &63.2\\
DSAN~\cite{DSAN} & 54.4 & 70.8 & 75.4 & 60.4 & 67.8 & 68.0 & 62.6 & 55.9 & 78.5 & 73.8 & 60.6 & 83.1 & 67.6\\
DCAN(ours)  & {\bfseries 58.0} & {\bfseries 76.2} & {\bfseries 79.3} & {\bfseries 67.3} & {\bfseries 76.1} & {\bfseries 75.6} & {\bfseries 65.4} & {\bfseries 56.0} & {\bfseries 80.7} & {\bfseries 74.2} & {\bfseries 61.2} & {\bfseries 84.2} & {\bfseries 71.2}\\
  & $\pm$0.3 & $\pm$0.1 & $\pm$0.2 & $\pm$0.2 & $\pm$0.3 & $\pm$0.2 & $\pm$0.1 & $\pm$0.2 & $\pm$0.1 & $\pm$0.3 & $\pm$0.6 & $\pm$0.3 & \\
\hline
\end{tabular}
\end{table*}

\begin{table*}[htb]
\centering
\renewcommand{\tabcolsep}{0.5pc} 
\renewcommand{\arraystretch}{1.0} 
\caption{Comparison with state-of-the-art methods on VISDA-2017 for unsupervised domain adaptation (ResNet-101).}
\label{table3}
\begin{tabular}{cccccccccccccc}
\hline
Method      &  plane & bcycl & bus & car & horse & knife & mcycl & person & plant & sktbrd & train & truck &  mean\\
\hline
ResNet~\cite{he2016deep}   & 72.3 & 6.1 & 63.4 & {\bfseries 91.7} & 52.7 & 7.9 & 80.1 & 5.6 & 90.1 & 18.5 & 78.1 & 25.9 & 49.4\\
DANN~\cite{ganin2016domain} & 81.9 & 77.7 & 82.8 & 44.3 & 81.2 & 29.5 & 65.1 & 28.6 & 51.9 & 54.6 & 82.8 & 7.8 & 57.4 \\
DAN~\cite{long2015learning} & 68.1 & 15.4 & 76.5 & 87.0 & 71.1 & 48.9 & 82.3 & 51.5 & 88.7 & 33.2 & 88.9 & 42.2 & 62.8\\
JAN~\cite{long2017deep} & 75.7 & 18.7 & 82.3 & 86.3 & 70.2 & 56.9 & 80.5 & 53.8 & 92.5 & 32.2 & 84.5 & {\bfseries 54.5} & 65.7 \\
MCD~\cite{saito2018maximum}  & 87.0 & 60.9 & 83.7 & 64.0 & 88.9 & 79.6 & 84.7 & 76.9 & 88.6 & 40.3 & 83.0 & 25.8 & 71.9 \\
SAFN~\cite{xu2018unsupervised} & 93.6 & 61.3 & {\bfseries 84.1} & 70.6 & {\bfseries 94.1} & 79.0 & 91.8 & {\bfseries 79.6} & 89.9 & 55.6 & 89.0 & 24.4 & 76.1 \\
DSAN~\cite{DSAN} & 90.9 & 66.9 & 75.7 & 62.4 & 88.9 & 77.0 & {\bfseries 93.7} & 75.1 & {\bfseries 92.8} & 67.6 & {\bfseries 89.1} & 39.4 & 75.1 \\
DCAN(ours) & {\bfseries 94.9} &	{\bfseries 83.7} &	75.7 &	56.5 &	92.9 &	{\bfseries 86.8} &	83.8 &	76.5 &	88.4 &	{\bfseries 81.6} &	84.2 &	51.1 &	{\bfseries 79.7}\\
\hline
\end{tabular}
\end{table*}

\subsection{Experiment Results on UDA Tasks}

In this section, we report the classification performance of DCAN on the UDA tasks, and compare it with several state-of-the-art methods, e.g., DAN~\cite{long2015learning}, BDA~\cite{wang2017balanced}, DANN~\cite{ganin2016domain}, Deep Reconstruction-Classification Network (DRCN)~\cite{Ghifary2016Deep}, Coupled Generative Adversarial Network (CoGAN)~\cite{Liu2016Coupled}, Generate To Adapt (GTA)~\cite{Sankaranarayanan2017Generate}, RTN~\cite{long2016unsupervised}, ADDA~\cite{tzeng2017adversarial}, JAN~\cite{long2017deep}, MADA~\cite{pei2018multi}, WDAN~\cite{yan2017mind}, CDAN~\cite{long2018conditional}, MCD~\cite{saito2018maximum}, SAFN~\cite{xu2018unsupervised}, GAACN~\cite{GAACN}, Jigsaw~\cite{Jigsaw}, Geometry-aware Dual-stream Network (GDN)~\cite{YANG2021107638}, DSAN~\cite{DSAN}. For a fair comparison, the same network protocol is used to evaluate the performance of BDA~\cite{wang2017balanced} and WDAN~\cite{yan2017mind}. For the rest methods, the reported results are cited from the original papers or from~\cite{DSAN}.

We first evaluate DCAN on three tasks of the digital datasets, i.e., SVHN$\rightarrow$MNIST, MNIST$\rightarrow$USPS and USPS$\rightarrow$MNIST. The results are shown in Table~\ref{table1}. We can see that DCAN achieves state-of-the-art results in all three tasks. In particular, the average classification accuracies of DCAN on these tasks are 97.5\%, 98.5\% and 98.7\%, which outperform the second best results by 0.6\%, 0.1\% and 4.1\%.

Table \ref{table2} shows the results of DCAN on six tasks of Office-31 dataset. We can see that DCAN achieves comparable performance on all of these tasks. DCAN achieves the average classification accuracy over all six tasks of 88.7\%, which is 1.0\% and 0.3\% better than the two latest conditional distribution alignment-based UDA methods, CDAN and DSAN, respectively. It implies that CMMD can better align the conditional distributions of source and target domains.

Office-Home contains 65 different categories, and the category distribution discrepancy across different domains are even larger, thus it provides a more challenging UDA benchmark. Table \ref{table4} shows classification accuracies of DCAN and the compared methods on all 12 transfer tasks. We obtain the following observations from Table \ref{table4}: 1) Marginal distribution alignment-based methods, such as DAN and DANN, perform poorly in Office-Home, and it can be attributed to the negative migration phenomenon. 2) CADN uses Conditional GAN to align feature spaces and achieves better performance than DANN and ADDA, which verifies the role of conditional distribution matching in UDA tasks. 3) Our DCAN uses CMMD to directly measure the discrepancy between two conditional distributions, which is more effective than previous conditional distribution alignment-based UDA methods. Experiment results validate the effectiveness of our method. As we can see from Table \ref{table4}, the average accuracy of DCAN is 71.2\%, which is at least 3.6\% higher than the compared approaches. It further validates the effectiveness of our method.

VisDA-2017 dataset is widely used to evaluate the classification performance of UDA methods on large-scale data. Table~\ref{table3} shows the experiment results of task synthetic$\rightarrow$real on VisDA-2017 dataset. It can be seen that DCAN achieves competitive performance with the state-of-the-art methods in most categories, and achieves the highest average classification accuracy of 79.7\%. Compared with the average classification accuracy of DSAN, DCAN has more than 4\% increments, which benefits from the ability of the discriminant information extraction module in DCAN to better mine discriminant information in the large-scale target domain dataset.

\subsection{Experiment Results on Partial UDA Tasks}

\begin{table*}[htb]
\centering
\renewcommand{\tabcolsep}{1pc} 
\renewcommand{\arraystretch}{1.0} 
\caption{Comparison with state-of-the-art methods on Office-31 for Partial unsupervised domain adaptation (ResNet-50).}
\label{table5}
\begin{tabular}{cccccccc}
\hline
Method      &  A$\rightarrow$W & D$\rightarrow$W & W$\rightarrow$D & A$\rightarrow$D & D$\rightarrow$A & W$\rightarrow$A & Avg\\
\hline
Resnet-50~\cite{he2016deep}   & 54.5 & 94.6 & 94.3 & 65.6 & 73.2 & 71.7 & 75.6\\
DAN~\cite{long2015learning}      	& 46.4 & 53.6 & 58.6 & 42.7 & 65.7 & 65.3 & 55.4\\
DANN~\cite{ganin2016domain}        & 41.4 & 46.8 & 38.9 & 41.4 & 41.3 & 44.7 & 42.4\\
RTN~\cite{long2016unsupervised}         & 75.3 & 97.1 & 98.3 & 66.9 & 85.6 & 85.7 & 84.8\\
JAN~\cite{long2017deep}       	& 43.4 & 53.6 & 41.4 & 35.7 & 51.0 & 51.6 & 46.1\\
BDA~\cite{wang2017balanced}       	& 77.6 & 95.6 & 98.7 & 82.2 & 86.5 & 86.4 & 87.8\\
PADA~\cite{cao2018partial}        & 86.5 & 99.3 & {\bfseries 100} & 82.2 & 92.7 & 95.4 & 92.7\\
WDAN~\cite{yan2017mind}        & 85.1 & 98.9 & {\bfseries 100} & 84.7 & 94.0 & 92.8 & 92.6\\
ETN~\cite{cao2019learning} & 94.5$\pm$0.2 & \textbf{100}$\pm$0.0 & \textbf{100}$\pm$0.0 & 95.0$\pm$0.2 & \textbf{96.2}$\pm$0.3 & 94.6$\pm$0.2 & 96.7\\
RTNet~\cite{chen2020selective} & 95.1$\pm$0.3 & \textbf{100}$\pm$0.0 & \textbf{100}$\pm$0.0 & \textbf{97.8}$\pm$0.1 & 93.9$\pm$0.1 & 94.1$\pm$0.1 & \textbf{96.8}\\
Jigsaw~\cite{Jigsaw} &91.8& 94.1& 98.9& 90.9& 89.9 &93.4 &93.2\\
DCAN(ours)  & \textbf{95.4}$\pm$0.5 & \textbf{100}$\pm$0.0 & \textbf{100}$\pm$0.0 &93.8$\pm$0.3 &95.8$\pm$0.2 & {\bfseries 95.8}$\pm$0.1 &\textbf{96.8}\\
\hline
\end{tabular}
\end{table*}

\begin{table*}[htb]
\footnotesize
\centering
\renewcommand{\tabcolsep}{0.21pc} 
\renewcommand{\arraystretch}{1.0} 
\caption{Comparison with state-of-the-art methods on Office-Home for Partial unsupervised domain adaptation (ResNet-50).}
\label{table6}
\begin{tabular}{cccccccccccccc}
\hline
Method      &  A$\rightarrow$C & A$\rightarrow$P & A$\rightarrow$R & C$\rightarrow$A & C$\rightarrow$P & C$\rightarrow$R & P$\rightarrow$A & P$\rightarrow$C & P$\rightarrow$R & R$\rightarrow$A & R$\rightarrow$C & R$\rightarrow$P & Avg\\
\hline
Resnet-50~\cite{he2016deep}   & 38.6 & 60.8 & 75.2 & 40.0 & 48.1 & 52.9 & 49.7 & 30.9 & 70.8 & 65.4 & 41.8 & 70.4 & 53.7 \\
DAN~\cite{long2015learning}       	& 44.4 & 61.8 & 74.5 & 41.8 & 45.2 & 54.1 & 46.9 & 38.1 & 68.4 & 64.4 & 45.4 & 68.9 & 54.5 \\
DANN~\cite{ganin2016domain}       	& 44.9 & 54.1 & 69.0 & 36.3 & 34.3 & 45.2 & 44.1 & 38.0 & 68.7 & 53.0 & 34.7 & 46.5 & 47.4  \\
RTN~\cite{long2016unsupervised}       	& 49.4 & 64.3 & 76.2 & 47.6 & 51.7 & 57.7& 50.4 & 41.5 & 75.5 & 70.2 & 51.8 & 74.8 & 59.3 \\
BDA~\cite{wang2017balanced}        	& 46.2 & 63.5 & 73.0 & 55.7 & 55.7 & 63.2 & 52.4 & 41.7 & 71.5 & 59.2 & 50.2 & 73.7 & 58.8 \\
PADA~\cite{cao2018partial}     & 52.0 & 67.0 & 78.7 & 52.2 & 53.8 & 59.0 & 52.6 & 43.2 & 78.8 & 73.7 & 56.6 & 77.1 & 62.1\\
WDAN~\cite{yan2017mind}     & 52.4 & 74.5 & 81.3 & 63.7 & 66.1 & 72.4 & 63.1 & 51.3 & 80.3 & 73.0 & 56.2 & 80.0 & 67.9\\
ETN~\cite{cao2019learning}          & 59.2 & 77.0 & 79.5 & 62.9 & 65.7 & 75.0 & 68.3 & 55.4 & 84.4 & 75.7 & 57.7 & 84.5 & 70.5\\
SAFN~\cite{xu2018unsupervised}      & 58.9$\pm$0.5 & 76.3$\pm$0.3 & 81.4$\pm$0.3 & 70.4$\pm$0.5 & 73.0$\pm$1.4 & 77.8$\pm$0.5 & {\bfseries 72.4}$\pm$0.3 & 55.3$\pm$0.5 & 80.4$\pm$0.8 & 75.8$\pm$0.4 & 60.4$\pm$0.8 & 79.9$\pm$0.2 & 71.8 \\
RTNet~\cite{chen2020selective} & {\bfseries 62.7}$\pm$0.1 & 79.3$\pm$0.2 & 81.2$\pm$0.1 & 65.1$\pm$0.1 & 68.4$\pm$0.3 & 76.5$\pm$0.1 & 70.8$\pm$0.2 & 55.3$\pm$0.1 & 85.2 $\pm$0.3& 76.9$\pm$0.2 & 59.1$\pm$0.2 & 83.4$\pm$0.3 & 72.0\\
DCAN(ours)  & 60.2$\pm$0.1 & {\bfseries 85.3}$\pm$0.2 & {\bfseries 89.0}$\pm$0.1 & {\bfseries 73.8}$\pm$0.4 & {\bfseries 76.9} $\pm$1.0 & {\bfseries 83.6}$\pm$0.1 & 71.7$\pm$0.2 & {\bfseries 59.4}$\pm$0.6 & {\bfseries 88.0}$\pm$0.4 & {\bfseries 79.6}$\pm$0.2 & {\bfseries 63.3}$\pm$0.8 & {\bfseries 85.3}$\pm$0.4 & {\bfseries 76.3}\\
\hline
\end{tabular}
\end{table*}

In this section, we report the performance of DCAN on the partial UDA tasks, and compare it with several state-of-the-art methods, i.e. ResNet~\cite{he2016deep}, DAN~\cite{long2015learning}, DANN~\cite{ganin2016domain}, RTN~\cite{long2016unsupervised}, JAN~\cite{long2017deep}, BDA~\cite{wang2017balanced}, IWAN~\cite{zhang2018importance}, PADA~\cite{cao2018partial}, WDAN~\cite{yan2017mind}, SAFN~\cite{xu2018unsupervised}, ETN~\cite{cao2019learning}, RTNet~\cite{chen2020selective} and Jigsaw~\cite{Jigsaw}. For a fair comparison, the same network backbone is used to evaluate the performance of BDA~\cite{wang2017balanced} and WDAN~\cite{yan2017mind}. For the rest methods, the reported results are cited from the original papers or from~\cite{cao2018partial}.

\begin{figure*}[htb]
\subfigure[$\lambda_{0}$]{\label{Fig:zhexian1}
\begin{minipage}[b]{0.235\textwidth} 
\centering \scalebox{0.32}{ 
\includegraphics{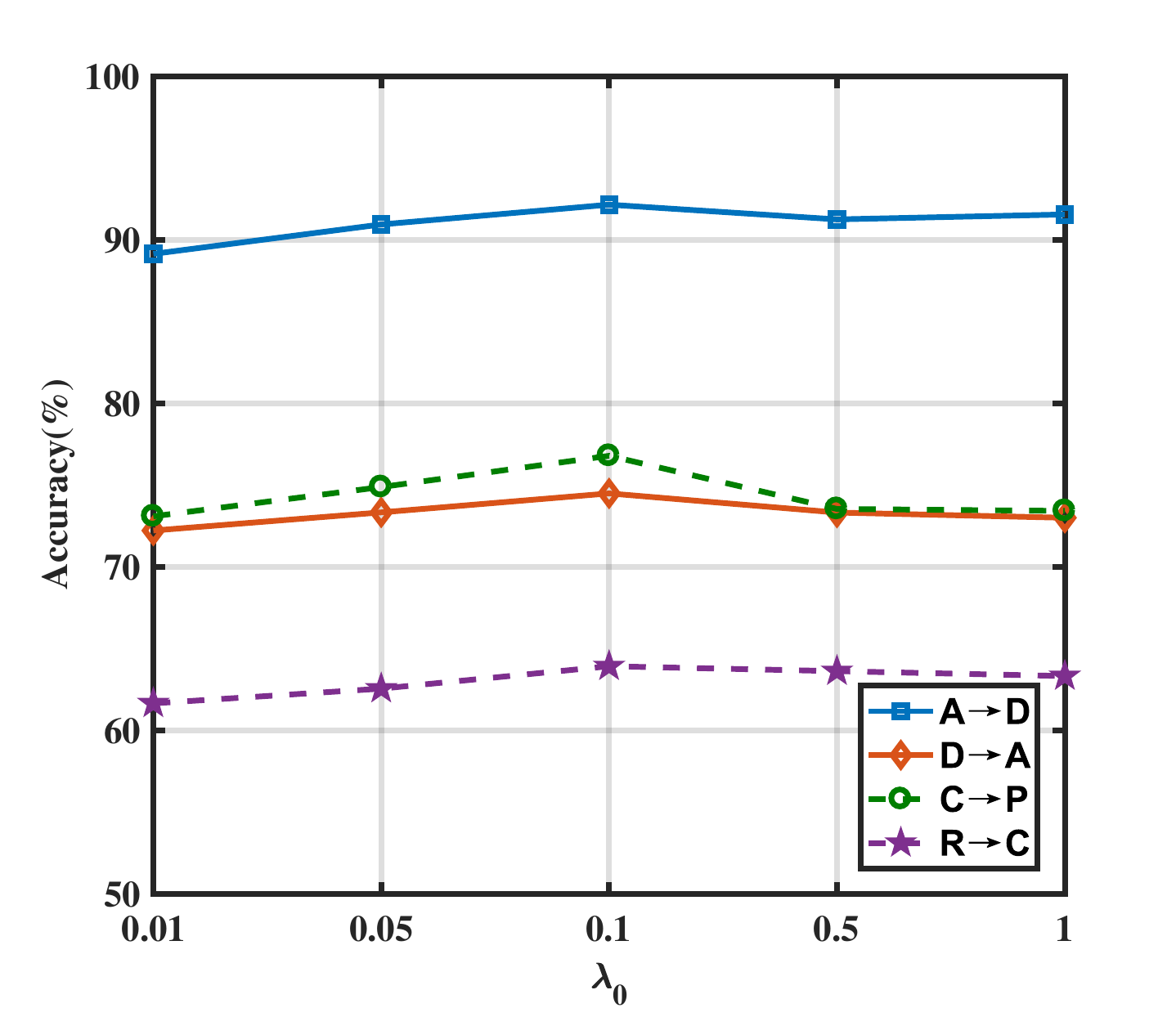}}
\end{minipage}}
\subfigure[$\lambda_{1}$]{\label{Fig:zhexian2}
\begin{minipage}[b]{0.235\textwidth}
\centering \scalebox{0.32}{ 
\includegraphics{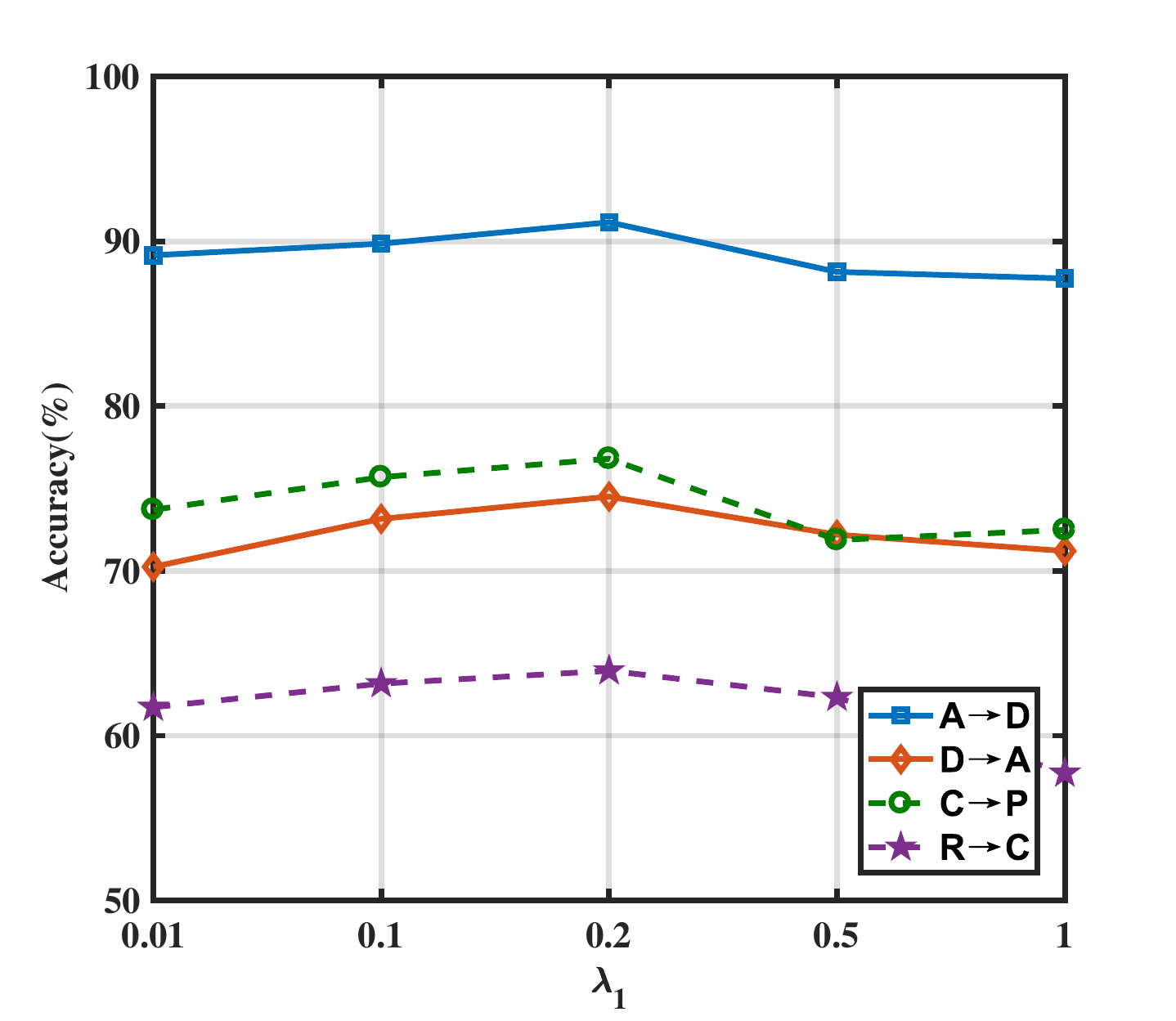}}
\end{minipage}}
\subfigure[$\gamma_{0}$]{\label{Fig:zhexian3}
\begin{minipage}[b]{0.235\textwidth}
\centering \scalebox{0.32}{ 
\includegraphics{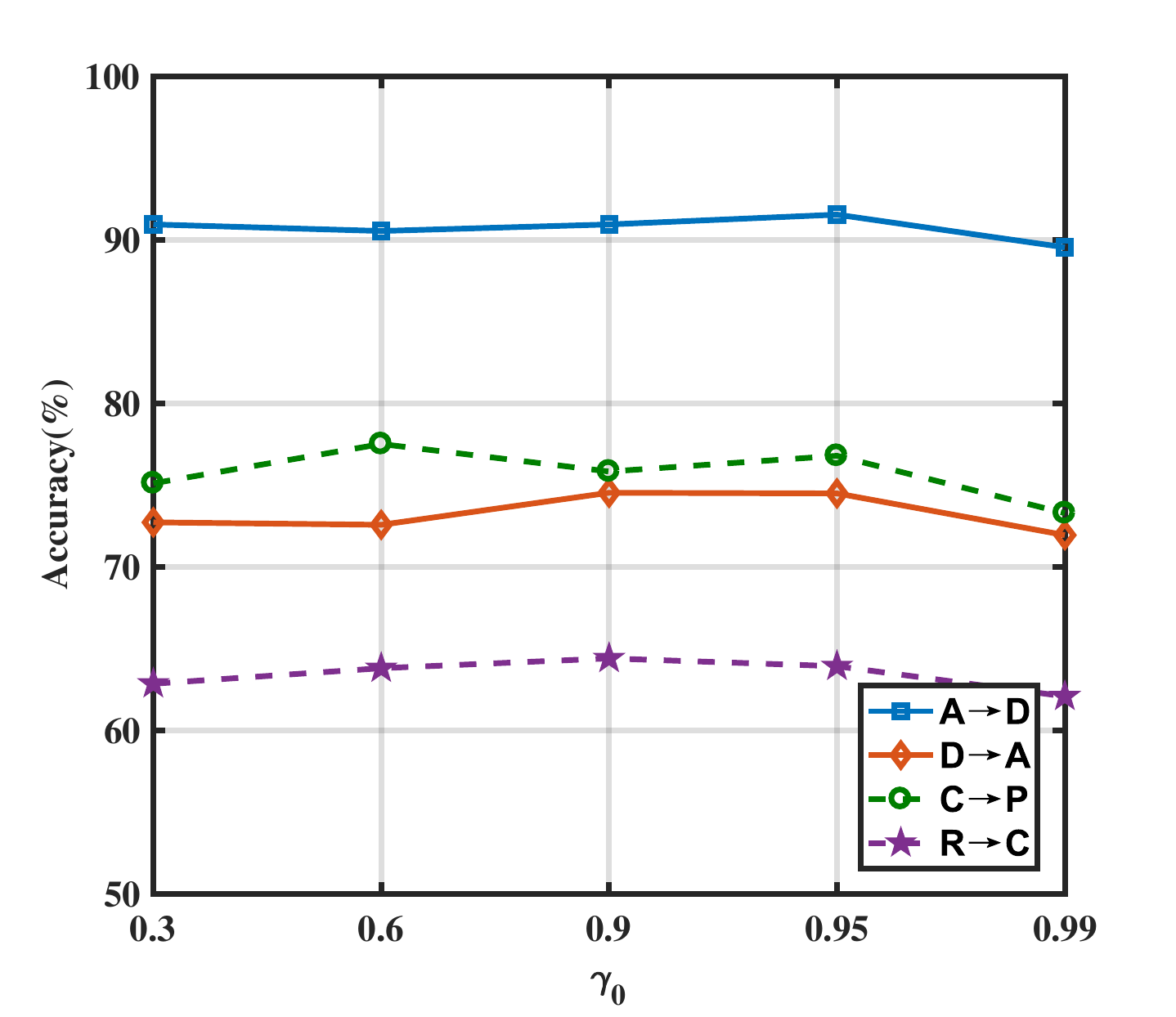}}
\end{minipage}}
\subfigure[$\gamma_{1}$]{\label{Fig:zhexian4}
\begin{minipage}[b]{0.235\textwidth}
\centering \scalebox{0.32}{ 
\includegraphics{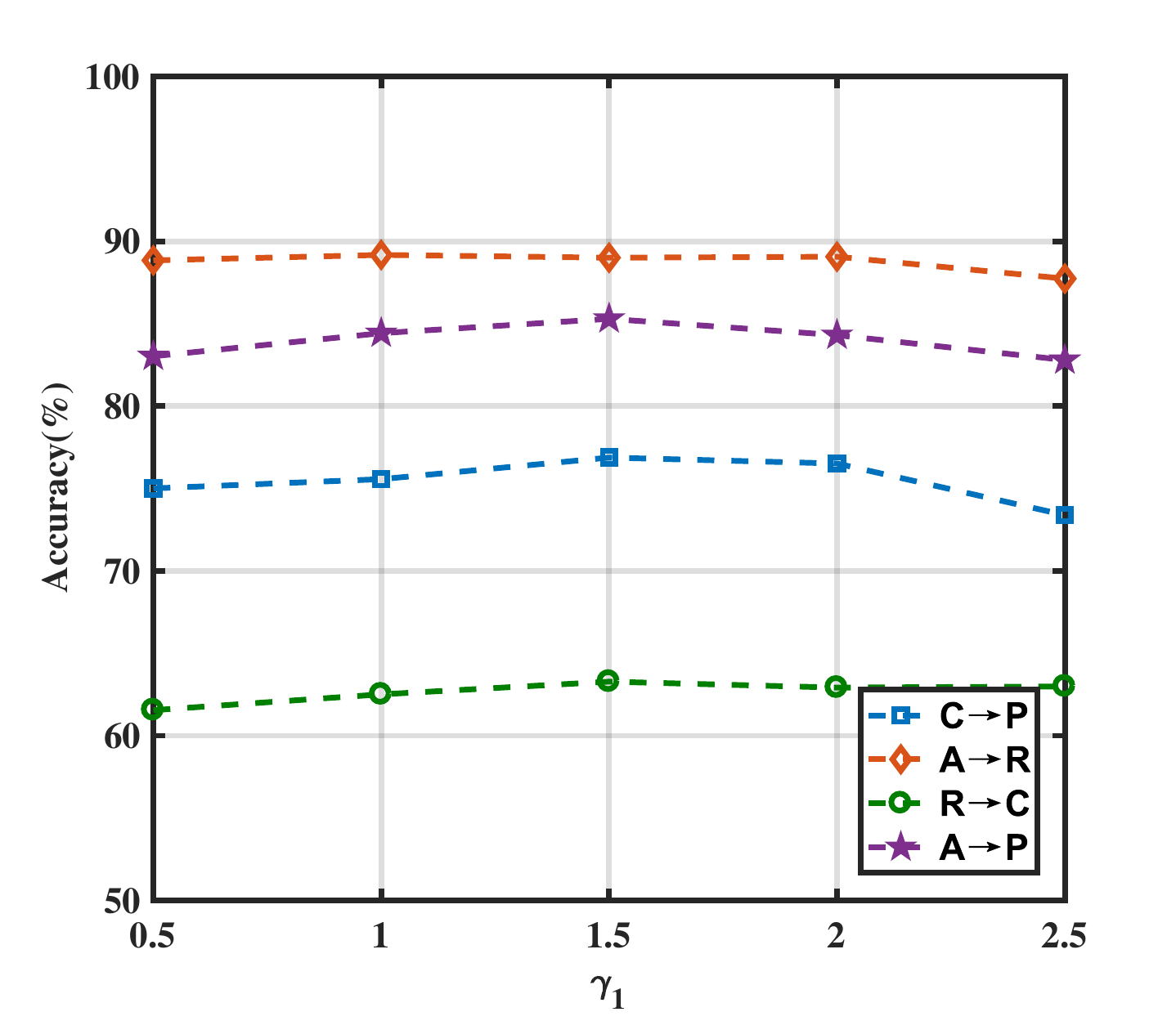}}
\end{minipage}}
\caption{(a)$\sim$(c) Sensitivity of DACN on UDA tasks and Partial UDA tasks to the hyper-parameter $\lambda_{0}$,$\lambda_{1}$,$\gamma_{0}$, respectively. (d) Sensitivity of DACN on Partial UDA tasks to the hyper-parameter $\gamma_{1}$. The solid lines represent the results on UDA tasks, and the dotted lines represent the results on Partial UDA tasks.}
\label{Fig:3}
\end{figure*}

Under the Partial UDA settings, the classification accuracy on 6 tasks of the Office-31 dataset and 12 tasks of the Office-Home dataset are shown in Tables \ref{table5} and \ref{table6}, respectively. We observe that the classification results obtained by marginal distribution alignment-based methods, such as DANN, ADDA and JAN, are even worse than directly finetune the ResNet-50. It indicates that marginal distribution alignment is seriously affected by negative transfer. Our DCAN uses conditional distributions, rather than marginal distributions, to achieve feature space alignment, thus it avoids the negative transfer effect. For Office-31, DCAN achieves the average classification accuracy over all six tasks of 96.8\%, which is the same as the state-of-the-art PDA method RTNet. On the more challenging Office-Home dataset, DCAN improves significantly compared to the state-of-the-art methods, and increase by 4.3\% in the average of all 12 tasks. These indicate that our new method can handle the Partial UDA Problems effectively.

\subsection{Effectiveness Analysis}

\begin{figure*}[htb]
\subfigure[Source-Only on Office-Home]{\label{Fig:home0}
\begin{minipage}[b]{0.32\textwidth} 
\centering \scalebox{0.35}{ 
\includegraphics{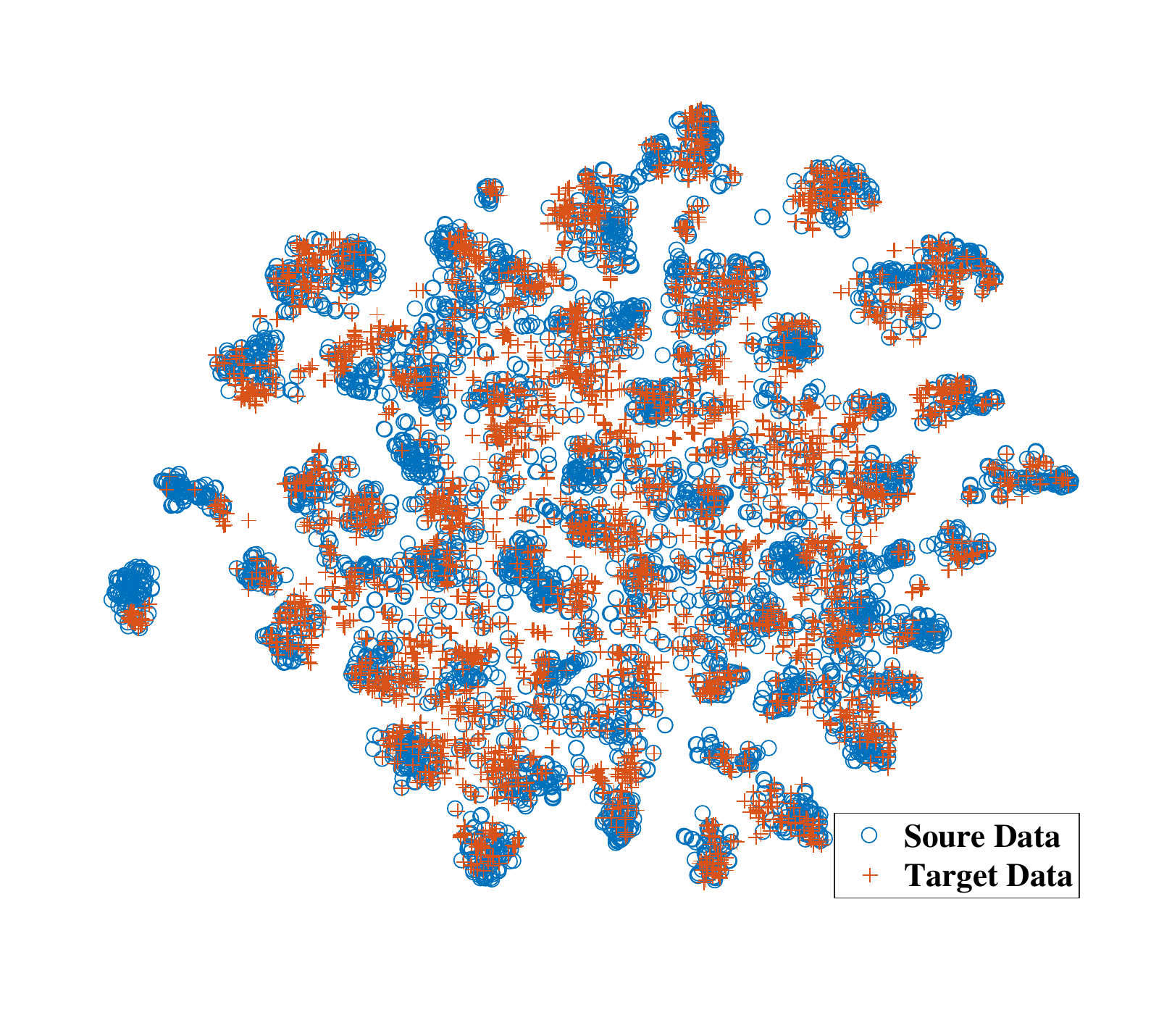}}
\end{minipage}}
\subfigure[DCAN on Office-Home]{\label{Fig:home2}
\begin{minipage}[b]{0.32\textwidth}
\centering \scalebox{0.35}{ 
\includegraphics{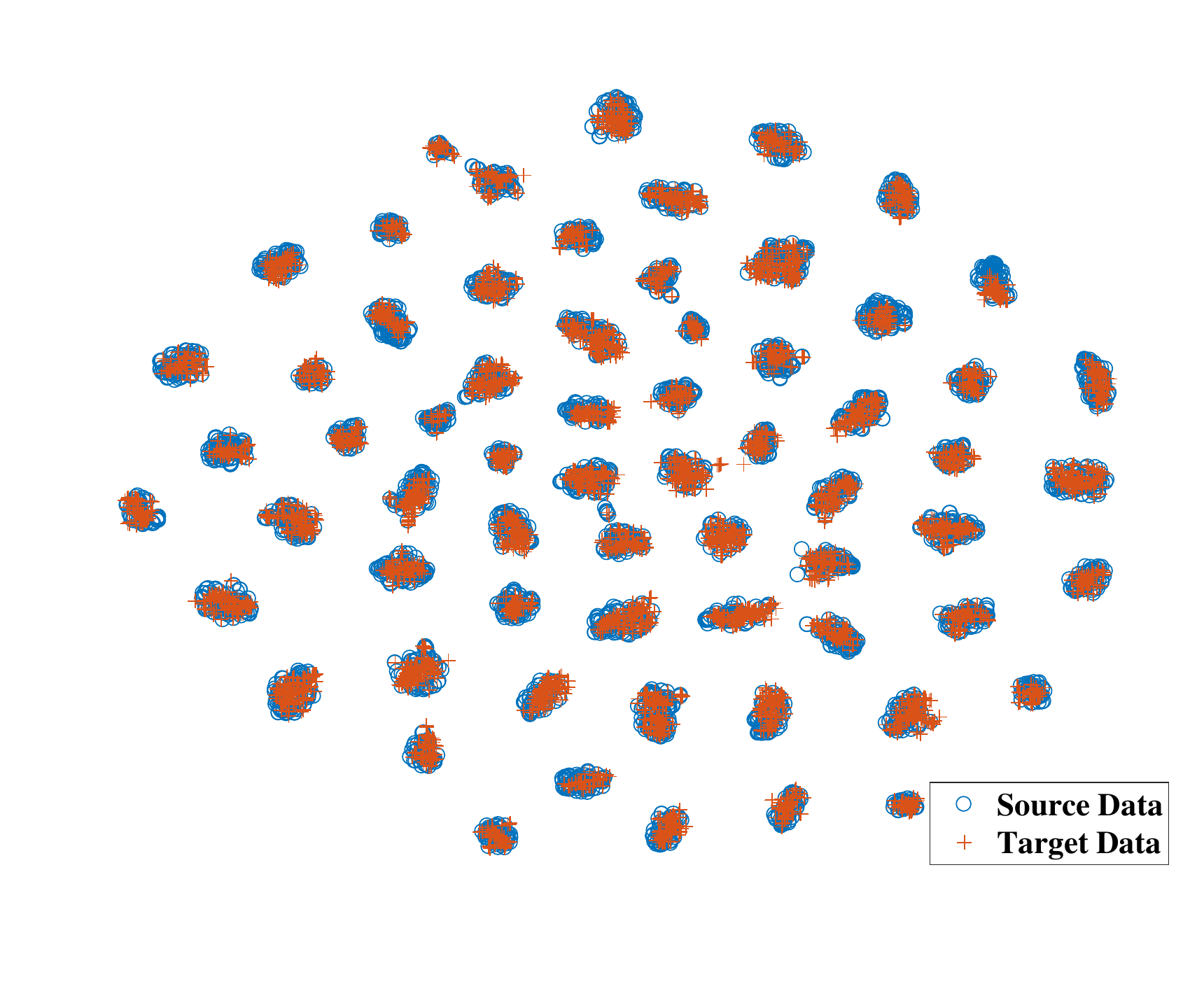}}
\end{minipage}}
\subfigure[DCAN on Office-Home under Partial setting]{\label{Fig:homep}
\begin{minipage}[b]{0.32\textwidth}
\centering \scalebox{0.35}{ 
\includegraphics{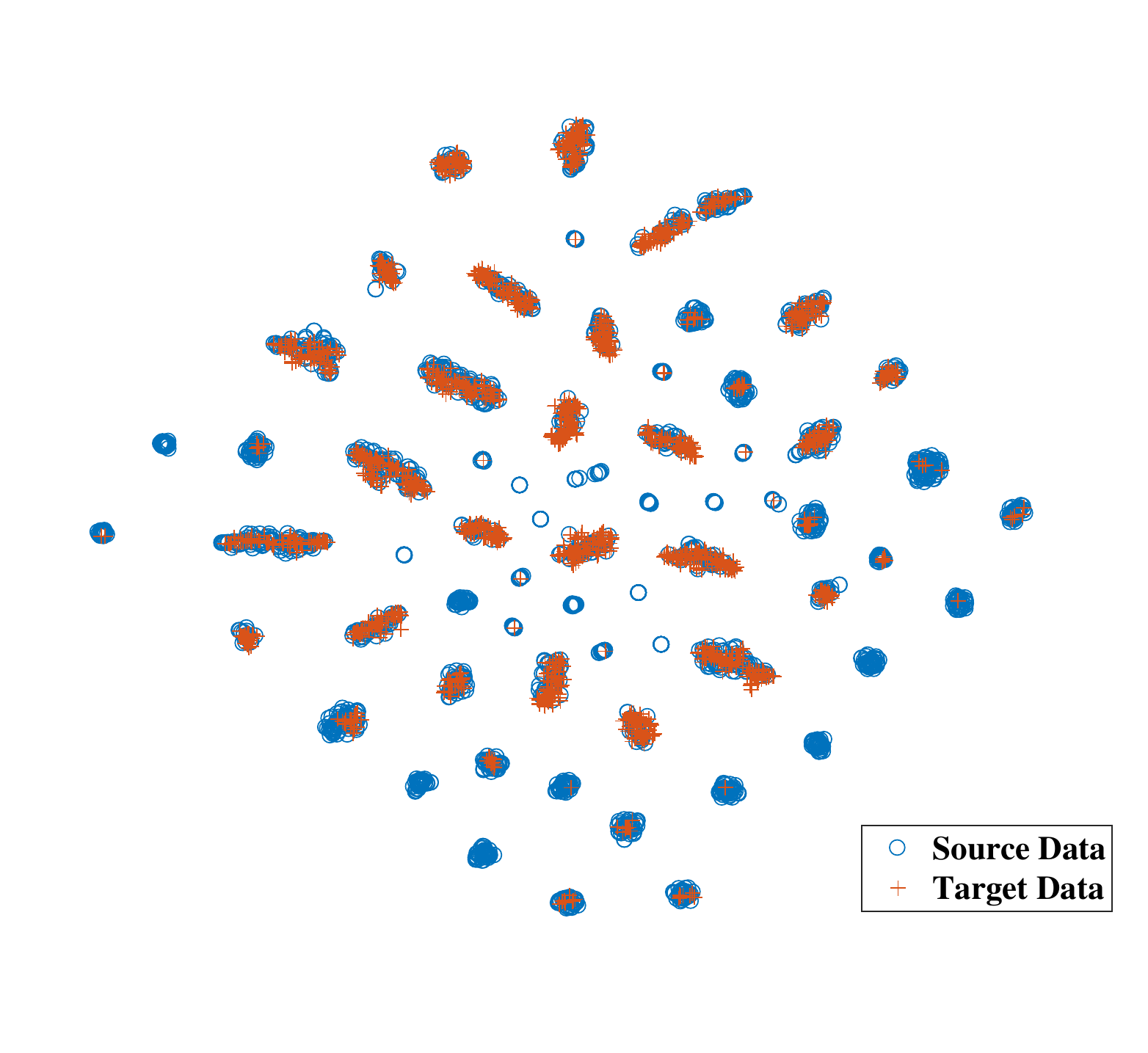}}
\end{minipage}}
\caption{T-SNE visualization in the task R$\rightarrow$P on Office-Home dataset. (a) Visualization of Source-Only model. (b) T-SNE visualization of our DCAN. (c) T-SNE visualization of our DCAN under partial setting. The circle and plus represent the source and target samples, respectively. Better viewed in color.}
\label{Fig:6}
\end{figure*}

\subsubsection{Parameter Sensitivity}\label{sect:Parameter Sensitivity}

There are four important hyper-parameters in DCAN, i.e., $\lambda _{0}$, $\lambda _{1}$, $\gamma_{0}$ and $\gamma_{1}$. To evaluate the sensitivity of CDAN against these hyper-parameters, we run CDAN with $\lambda_{0}\in \{0.01,0.05,0.1,0.5,1\}$, $\lambda_{1}\in \{0.01,0.1,0.2,0.5,1\}$, $\gamma_{0}\in \{0.3,0.6,0.9,0.95,0.99\}$, and $\gamma_{1}\in \{0.5,1,1.5,2,2.5\}$. We first investigate $\lambda_{0}$, $\lambda_{1}$ and $\gamma_{0}$ with the UDA tasks on Office-31 and the Partial UDA tasks on Office-Home. Then, we investigate $\gamma_{1}$ with the Partial UDA tasks on Office-Home. The experiment results are shown in Figure \ref{Fig:3}. From Figures \ref{Fig:zhexian1} and \ref{Fig:zhexian2}, we can see that the accuracy of DCAN has a similar trend for a majority of tasks when hyper-parameters $\lambda_{0}$ and $\lambda_{1}$ vary. The best choice of weight hyper-parameters $\lambda_{0}$, $\lambda _{1}$ are 0.1 and 0.2, respectively. $\gamma_{0}$ is the threshold for controlling pseudo-labels generation. A small threshold will reduce the accuracy of pseudo-labels, while a large threshold will reduce the number of pseudo-labels, thus a suitable threshold can neither be too large nor too small. The results shown in Figure~\ref{Fig:zhexian3} validate our motivation that the classification accuracy first increases and then decreases, and DCAN obtains the best results when $\gamma_{0} = 0.95$. $\gamma_{1}$ is used to control the entropy of $P(\hat{Y})$ in Partial DCAN. From Figure~\ref{Fig:zhexian4}, we observe that DCAN obtains a robust result against $\gamma_{1}$, and the optimal parameter is selected from $\{1,1.5,2\}$.

\subsubsection{Feature Visualization}

We use the t-SNE~\cite{Maaten2008Visualizing} method to visualize the embedding subspace of different methods in the R$\rightarrow$P task on the Office-Home dataset. The results are shown in Figure \ref{Fig:6}. As shown in Figure \ref{Fig:home0}, the Source-Only model cannot align feature spaces effectively because of the domain discrepancy. Figures \ref{Fig:home2} and \ref{Fig:homep} show the embedding space of DCAN and Partial DCAN, respectively. We obtain the following observations: 1) DCAN achieves conditional distribution alignment very well. 2) The inter-class scatter in the feature space of DCAN is larger. It indicates that DCAN can simultaneously achieve conditional distribution alignment and discriminant information extraction.

\subsubsection{Evaluation of Each Component}

\begin{figure}[htb]
\subfigure[Results on UDA tasks]{\label{Fig:zhexian5}
\begin{minipage}[b]{0.23\textwidth} 
\centering \scalebox{0.4}{ 
\includegraphics{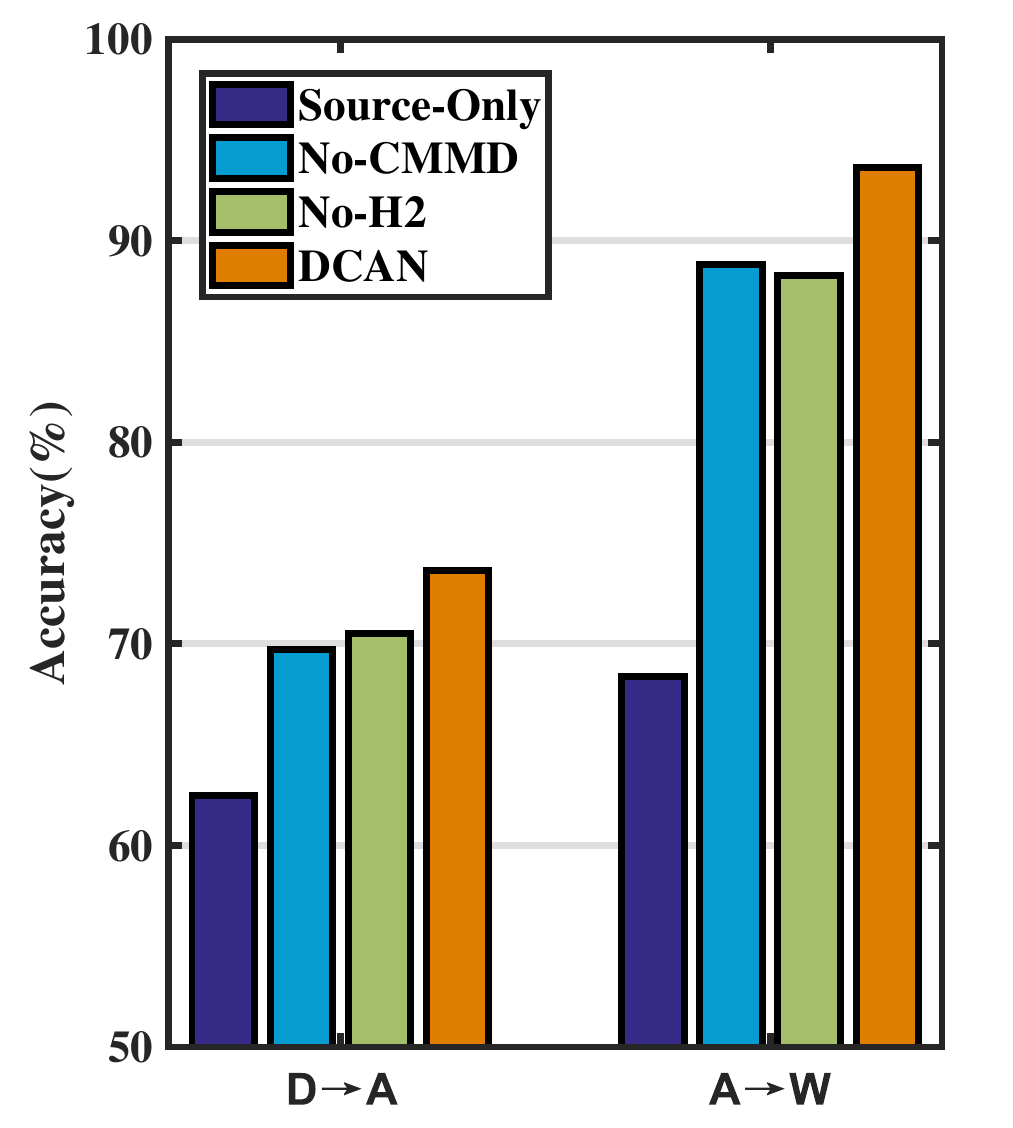}}
\end{minipage}}
\subfigure[Results on Partial UDA tasks]{\label{Fig:zhexian6}
\begin{minipage}[b]{0.23\textwidth}
\centering \scalebox{0.4}{ 
\includegraphics{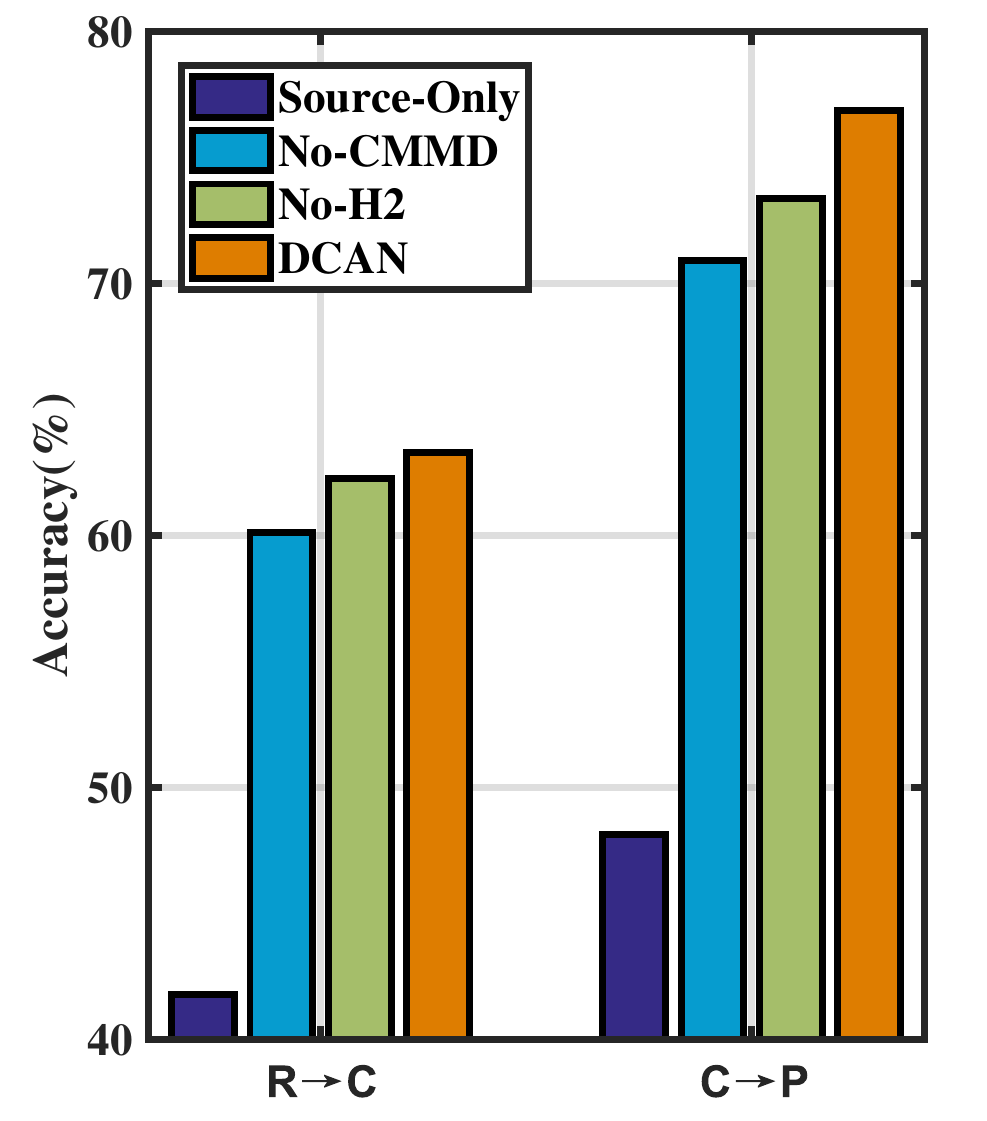}}
\end{minipage}}
\caption{(a) Evaluation of each component on UDA tasks. (b) Evaluation of each component on Partial UDA tasks.}
\label{fig:4}
\end{figure}

In order to achieve conditional distribution matching, DCAN contains three loss functions: 1) cross-entropy loss to extract discriminant information in the source domain, 2) CMMD loss to minimize the difference between conditional distributions, and 3) mutual information loss to extract target discrimination information from the target domain. Compared with other UDA methods, DCAN mainly introduces two new terms, i.e., CMMD and $H(P(\hat{Y}))$. Let No-CMMD and No-H2 denote the DCAN models without CMMD and $H(P(\hat{Y}))$, respectively. We evaluate the performance of these models on the UDA tasks and the Partial UDA tasks. Experiment results are shown in Figure~\ref{fig:4}. We observe that the performance of No-CMMD and No-H2 is better than Source-Only, but worse than DCAN. This indicates the importance of CMMD and $H(P(\hat{Y}))$ for improving classification performance in both UDA and Partial UDA tasks. In Figure~\ref{Fig:zhexian6}, we observe that the performance of No-CMMD is worse than No-H2 under the Partial UDA setting. This indicates that CMMD is more important than H(p($\hat{Y}$)) when the label distribution difference between source and target domains is large.

\subsubsection{Convergence Performance}

\begin{figure}[h]
\centering{\scalebox{0.55}{\includegraphics{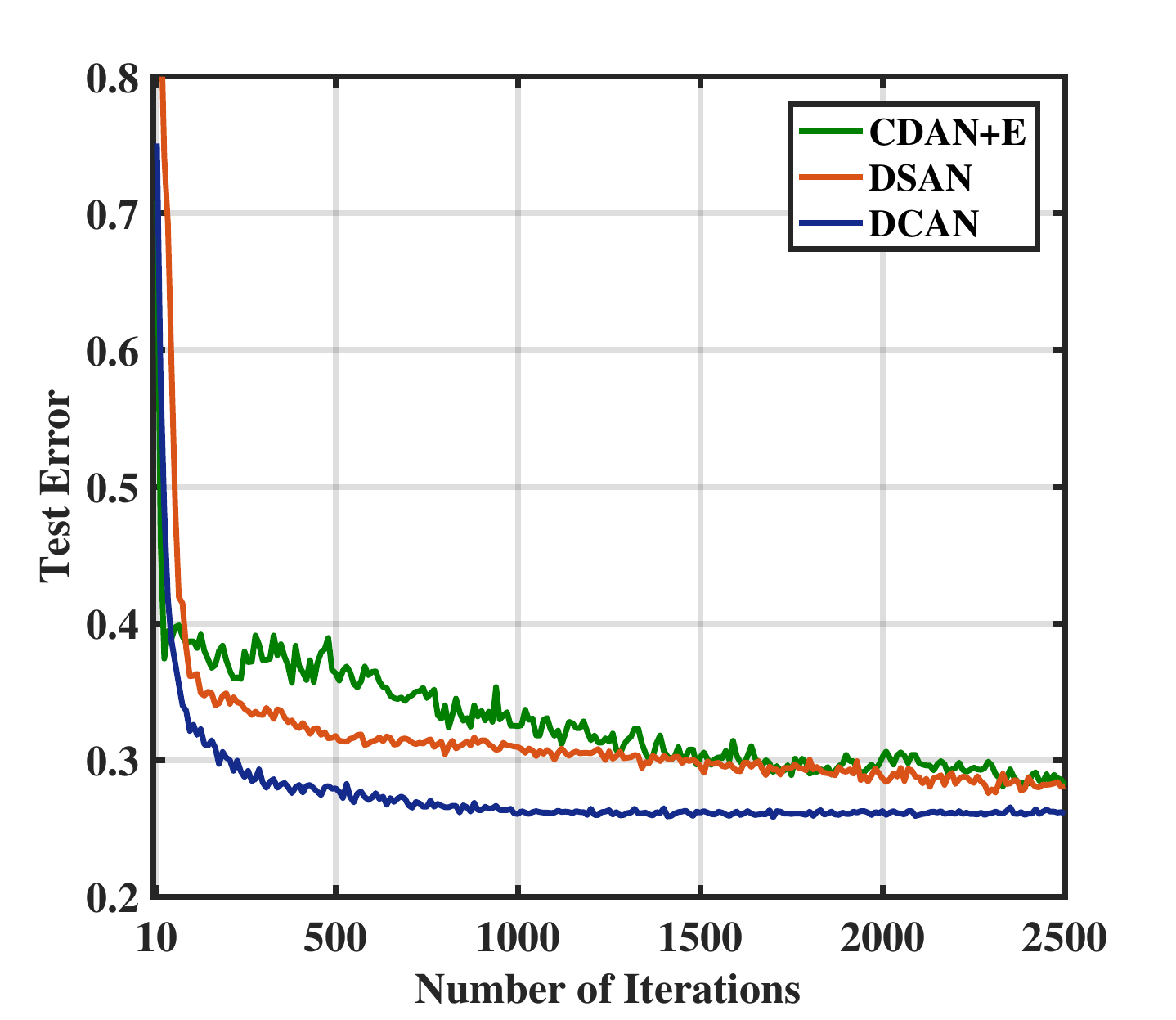}}\caption{Target test error w.r.t. the number of iterations.}\label{fig:test_error}} 
\end{figure}

We evaluate the convergence performance of three different conditional distribution alignment-based UDA methods, CDAN+E, DSAN and DCAN, with the target test errors of the UDA task D$\rightarrow$A on the Office-31 dataset. The results are shown in Figure~\ref{fig:test_error}. We can find that DCAN achieves faster convergence performance and lower test error than CDAN+E and DSAN. This indicates that DCAN can be trained more efficiently and stably than previous conditional distribution alignment-based UDA methods. There are two reasons why DCAN achieves better performance than previous conditional distribution alignment-based UDA methods. 1) Compared with class-specific MMD and conditional adversarial generation network, CMMD can estimate the conditional distribution discrepancy more efficiently. 2) DCAN introduces mutual information to extract the discriminant information in target domain, which effectively improves the accuracy of conditional distribution discrepancy estimation.

\subsubsection{Effects of Batch-size}

\begin{figure}[h]
\centering{\scalebox{0.55}{\includegraphics{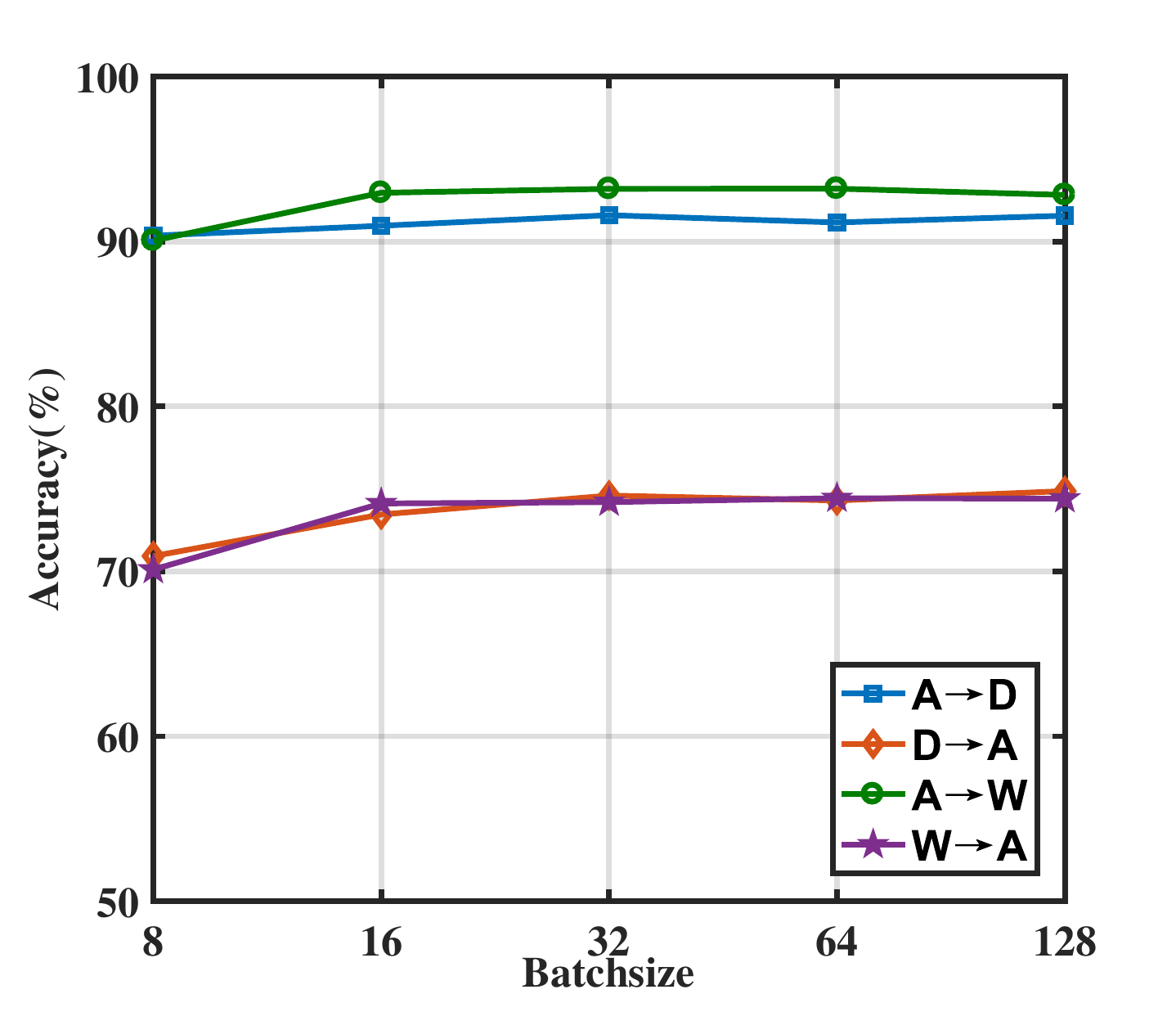}}
\caption{Sensitivity of DACN to the batch-size $n$ on Office-31.}\label{fig:bs}} 
\end{figure}
During the training process of DCAN, we need to estimate CMMD and mutual information on a mini-batch dataset. The accuracy of these estimators depends on the batch-size $n$. To test the effects of the setting of batch-size, we run DCAN with different batch-sizes $n\in \{8, 16, 32, 64, 128\}$. The experiment results are shown in Figure~\ref{fig:bs}. It can be observed that DCAN achieves a robust classification performance with regard to a wide range of $n$ values, and the classification performance of DCAN is degenerated when the value of batch-size is particularly small, such as $n=8$. This demonstrates the robustness of DCAN under the batch-size $n$. It also indicates that we do not need to deliberately fine-tune the batch-size in real implementations.

\subsubsection{DCAN with Ground Truth Labels}

\begin{figure}[h]
\centering{\scalebox{0.68}{\includegraphics{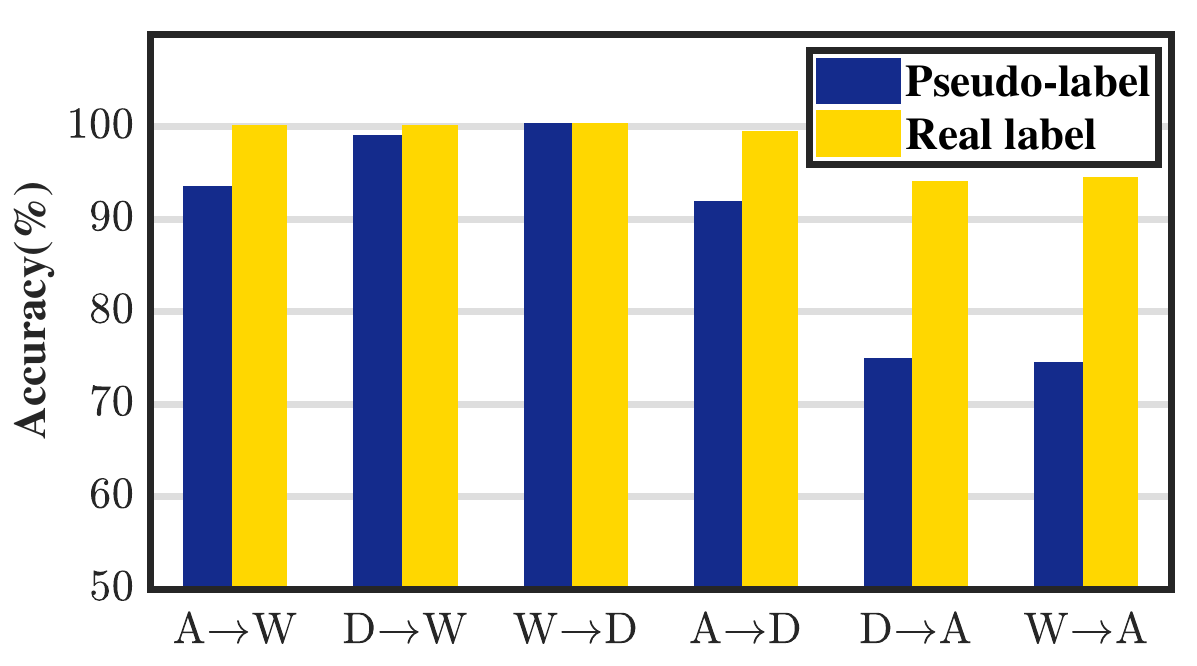}}
\caption{The classification performance of DCAN on Office-31 dataset when pseudo-labels and real labels are used to estimate CMMD, respectively.}\label{fig:r_label}} 
\end{figure}

\begin{figure}[h]
\centering{\scalebox{0.6}{\includegraphics{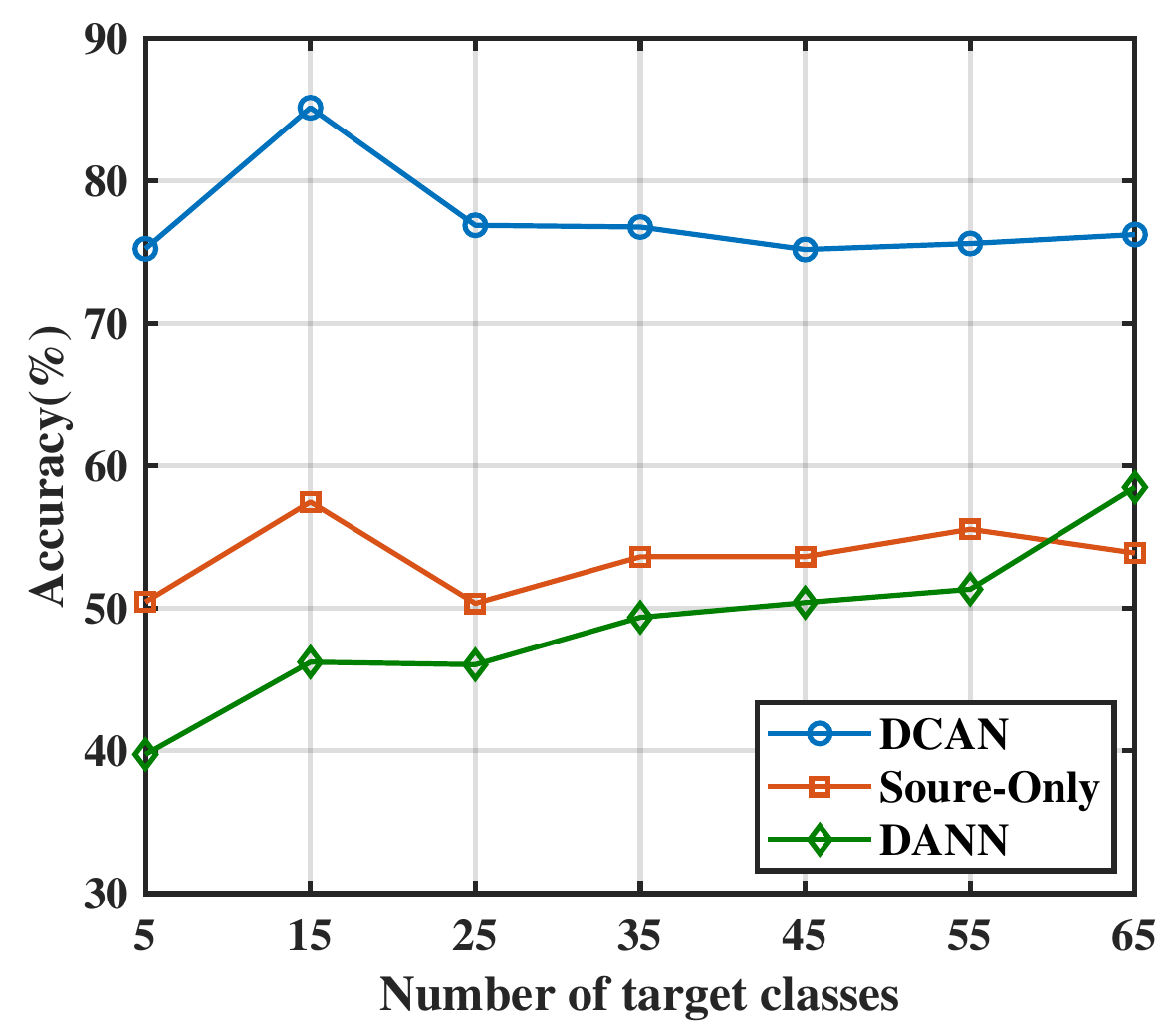}}\caption{The classification accuracy curve of different numbers of the target domain classes in the C$\rightarrow$P task on Office-Home dataset.}\label{fig:5}} 
\end{figure}

When estimating the CMMD between the conditional distributions, we replace the missing label information on the target domain with pseudo-labels, which affects the accuracy of the CMMD estimator. To further verify the effectiveness of DCAN, we evaluate the classification performance of DCAN on Office-31 when the ground truth labels of target domain are used to estimate CMMD during training. The results are shown in Figure~\ref{fig:r_label}. The classification performance of DCAN with real labels is greatly improved compared to DCAN with pseudo-labels. On A$\rightarrow$W and A$\rightarrow$D tasks, the classification accuracy of DCAN with real labels is close to 100$\%$. On more challenging W$\rightarrow$A and D$\rightarrow$A tasks, the classification accuracy of DCAN with real labels increases by about 20$\%$. These results indicate that DCAN can effectively transfer the classification information by aligning conditional distributions when the ground truth of target labels are given.

\subsubsection{Number of Classes in Target Domain}

We evaluate the impact of category number in the target domain. Figure~\ref{fig:5} shows the experiment results of three methods, i.e., Source-Only, DANN~\cite{ganin2016domain} and DCAN, in the C$\rightarrow$P task of Office-Home dataset. In this experiment, the source domain has 65 different categories, and the category number in the target domain is gradually reduced from 65 to 5. We see that as the category number decreases, the classification accuracy of DANN gradually decreases, even lower than the Source-Only model. Our DCAN shows robustness against the category number in the target domain, and achieves more than 75\% classification accuracy on all tasks. This indicates the effectiveness of DCAN on UDA tasks when the category distribution of $\mathcal{D}_s$ and $\mathcal{D}_t$ are different.

\begin{table}[h]
\centering
\renewcommand{\tabcolsep}{1.05pc} 
\renewcommand{\arraystretch}{1.15} 
\caption{Training Time (S) of DDC, DANN and DCAN.}
\label{table_time}
\begin{tabular}{ccccc}
\hline
Method      &  A$\rightarrow$W & D$\rightarrow$A & R$\rightarrow$C & C$\rightarrow$P\\
\hline
DDC     & 10.99 & 7.86 & 55.99 & 53.44\\
DANN  & 10.46 & 7.41 & 53.88 & 51.61\\
DCAN  & 11.05 & 8.16  & 58.06 & 54.60\\
\hline
\end{tabular}
\end{table}

\subsubsection{Time Complexity of DCAN}

To show the computational complexity of DCAN, we count the GPU time on four different tasks, i.e., A$\rightarrow$W, D$\rightarrow$A, R$\rightarrow$C and C$\rightarrow$P. The time is cost by one epoch, and it is compared with that of two widely used methods DDC~\cite{tzeng2014deep} and DANN\cite{ganin2016domain}. All experiments are run on a device with an NVIDIA GTX1080Ti GPU. From the experiment results as shown in Table~\ref{table_time}, we obtain the following observations: 1)~The training time of DCAN is about 6\% longer than DANN. The reason is that DCAN needs to calculate the kernel matrix $K$. The computational complexity for calculating $K$ is $O(n^{2}m)$, where $n$ is the batch-size and $m$ is the dimension of features. In practice, $m$ is usually large, which makes it time-consuming to calculate $K$. 2)~The training time of DANN is about 2\% longer than that of DDC. This is because the CMMD term needs to calculate the inverse of $K$. The computational complexity of this step is $O(n^{3})$. Fortunately, $n$ is usually small in deep learning, which means that it does not take much time to calculate the inverse of $K$.

\section{Conclusion}\label{sect:conclusion}		

This paper presents a novel domain adaptation method, DCAN, to learn conditional domain-invariant features for image classification. DCAN introduces the CMMD metric to achieve conditional distribution alignment of the source and target domains. Compared with the class-specific MMD and conditional adversarial generation network, CMMD can measure the conditional distribution discrepancy between source and target domains more efficiently. In addition, an extra mutual information is introduced to extract discriminant information from the target domain. DCAN considers the conditional dependence between features and labels when aligning feature spaces, thus it has wider application scenarios, such as Partial UDA. DCAN is evaluated on four benchmark databases, and the results show that it achieves state-of-the-arts performance in both UDA and Partial UDA tasks.

How to deal with more general problems such as Heterogeneous domain adaptation and zero-shot classification, are our future work.

\bibliographystyle{IEEEtran}
\bibliography{DCAN}

\begin{thebibliography}{10}
\providecommand{\url}[1]{#1}
\csname url@samestyle\endcsname
\providecommand{\newblock}{\relax}
\providecommand{\bibinfo}[2]{#2}
\providecommand{\BIBentrySTDinterwordspacing}{\spaceskip=0pt\relax}
\providecommand{\BIBentryALTinterwordstretchfactor}{4}
\providecommand{\BIBentryALTinterwordspacing}{\spaceskip=\fontdimen2\font plus
\BIBentryALTinterwordstretchfactor\fontdimen3\font minus
  \fontdimen4\font\relax}
\providecommand{\BIBforeignlanguage}[2]{{%
\expandafter\ifx\csname l@#1\endcsname\relax
\typeout{** WARNING: IEEEtran.bst: No hyphenation pattern has been}%
\typeout{** loaded for the language `#1'. Using the pattern for}%
\typeout{** the default language instead.}%
\else
\language=\csname l@#1\endcsname
\fi
#2}}
\providecommand{\BIBdecl}{\relax}
\BIBdecl

\bibitem{Krizhevsky2012ImageNet}
A.~Krizhevsky, I.~Sutskever, and G.~E. Hinton, ``Imagenet classification with
  deep convolutional neural networks,'' in \emph{Advances in Neural Information
  Processing Systems}, 2012, pp. 1097--1105.

\bibitem{chan2015pcanet}
T.-H. Chan, K.~Jia, S.~Gao, J.~Lu, Z.~Zeng, and Y.~Ma, ``Pcanet: A simple deep
  learning baseline for image classification?'' \emph{IEEE Transactions on
  Image Processing}, vol.~24, no.~12, pp. 5017--5032, 2015.

\bibitem{Donahue2014DeCAF}
J.~Donahue, Y.~Jia, O.~Vinyals, J.~Hoffman, Z.~Ning, E.~Tzeng, and T.~Darrell,
  ``Decaf: a deep convolutional activation feature for generic visual
  recognition,'' in \emph{International Conference on Machine Learning}, 2014,
  pp. 647--655.

\bibitem{Yosinski2014How}
J.~Yosinski, J.~Clune, Y.~Bengio, and H.~Lipson, ``How transferable are
  features in deep neural networks?'' in \emph{Advances in Neural Information
  Processing Systems}, 2014, pp. 3320--3328.

\bibitem{Oquab2014Learning}
M.~Oquab, L.~Bottou, I.~Laptev, and J.~Sivic, ``Learning and transferring
  mid-level image representations using convolutional neural networks,'' in
  \emph{IEEE Conference on Computer Vision and Pattern Recognition}, 2014, pp.
  1717--1724.

\bibitem{Pan2010A}
S.~J. Pan and Q.~Yang, ``A survey on transfer learning,'' \emph{IEEE
  Transactions on Knowledge and Data Engineering}, vol.~22, no.~10, pp.
  1345--1359, 2010.

\bibitem{Zhang2013Domain}
K.~Zhang, B.~Sch{\"o}lkopf, K.~Muandet, and Z.~Wang, ``Domain adaptation under
  target and conditional shift,'' in \emph{International Conference on Machine
  Learning}, 2013, pp. 819--827.

\bibitem{cao2018partialtransfer}
Z.~Cao, M.~Long, J.~Wang, and M.~I. Jordan, ``Partial transfer learning with
  selective adversarial networks,'' in \emph{IEEE Conference on Computer Vision
  and Pattern Recognition}, 2018, pp. 2724--2732.

\bibitem{cao2018partial}
Z.~Cao, L.~Ma, M.~Long, and J.~Wang, ``Partial adversarial domain adaptation,''
  in \emph{European Conference on Computer Vision}, 2018, pp. 135--150.

\bibitem{Donahue2013Semi}
J.~Donahue, J.~Hoffman, E.~Rodner, K.~Saenko, and T.~Darrell, ``Semi-supervised
  domain adaptation with instance constraints,'' in \emph{IEEE Conference on
  Computer Vision and Pattern Recognition}, 2013, pp. 1--8.

\bibitem{Ren2014Transfer}
C.~X. Ren, D.~Q. Dai, K.~K. Huang, and Z.~R. Lai, ``Transfer learning of
  structured representation for face recognition,'' \emph{IEEE Transactions on
  Image Processing}, vol.~23, no.~12, pp. 5440--5454, 2014.

\bibitem{Ren2018Generalized}
C.~X. Ren, X.~L. Xu, and H.~Yan, ``Generalized conditional domain adaptation: A
  causal perspective with low-rank translators,'' \emph{IEEE Transactions on
  Cybernetics}, vol.~50, no.~2, pp. 821--834, 2020.

\bibitem{tzeng2014deep}
E.~Tzeng, J.~Hoffman, N.~Zhang, K.~Saenko, and T.~Darrell, ``Deep domain
  confusion: Maximizing for domain invariance,'' \emph{arXiv preprint
  arXiv:1412.3474}, 2014.

\bibitem{long2015learning}
M.~Long, Y.~Cao, J.~Wang, and M.~Jordan, ``Learning transferable features with
  deep adaptation networks,'' in \emph{International Conference on Machine
  Learning}, 2015, pp. 97--105.

\bibitem{ganin2016domain}
Y.~Ganin, E.~Ustinova, H.~Ajakan, P.~Germain, H.~Larochelle, F.~Laviolette,
  M.~Marchand, and V.~Lempitsky, ``Domain-adversarial training of neural
  networks,'' \emph{Journal of Machine Learning Research}, vol.~17, pp.
  2096--2030, 2016.

\bibitem{tzeng2017adversarial}
E.~Tzeng, J.~Hoffman, K.~Saenko, and T.~Darrell, ``Adversarial discriminative
  domain adaptation,'' in \emph{IEEE Conference on Computer Vision and Pattern
  Recognition}, 2017, pp. 7167--7176.

\bibitem{long2013transfer}
M.~Long, J.~Wang, G.~Ding, J.~Sun, and P.~S. Yu, ``Transfer feature learning
  with joint distribution adaptation,'' in \emph{IEEE international conference
  on computer vision}, 2013, pp. 2200--2207.

\bibitem{wang2017balanced}
J.~Wang, Y.~Chen, S.~Hao, W.~Feng, and Z.~Shen, ``Balanced distribution
  adaptation for transfer learning,'' in \emph{IEEE International Conference on
  Data Mining}, 2017, pp. 1129--1134.

\bibitem{long2018conditional}
M.~Long, Z.~Cao, J.~Wang, and M.~I. Jordan, ``Conditional adversarial domain
  adaptation,'' in \emph{Advances in Neural Information Processing Systems},
  2018, pp. 1647--1657.

\bibitem{xu2018unsupervised}
R.~Xu, G.~Li, J.~Yang, and L.~Lin, ``Larger norm more transferable: An adaptive
  feature norm approach for unsupervised domain adaptation,'' in \emph{IEEE
  International Conference on Computer Vision}, 2019, pp. 1426--1435.

\bibitem{zhang2019category}
Q.~Zhang, J.~Zhang, W.~Liu, and D.~Tao, ``Category anchor-guided unsupervised
  domain adaptation for semantic segmentation,'' in \emph{Advances in Neural
  Information Processing Systems}, 2019, pp. 435--445.

\bibitem{saito2018maximum}
K.~Saito, K.~Watanabe, Y.~Ushiku, and T.~Harada, ``Maximum classifier
  discrepancy for unsupervised domain adaptation,'' in \emph{Proceedings of the
  IEEE Conference on Computer Vision and Pattern Recognition}, 2018, pp.
  3723--3732.

\bibitem{HUANG2022108384}
``Multi-level adversarial network for domain adaptive semantic segmentation,''
  \emph{Pattern Recognition}, vol. 123, p. 108384, 2022.

\bibitem{CASTELLANOS2021108099}
``Unsupervised neural domain adaptation for document image binarization,''
  \emph{Pattern Recognition}, vol. 119, p. 108099, 2021.

\bibitem{long2016unsupervised}
M.~Long, J.~Wang, and M.~I. Jordan, ``Unsupervised domain adaptation with
  residual transfer networks,'' in \emph{Advances in Neural Information
  Processing Systems}, 2016, pp. 136--144.

\bibitem{wang2018visual}
J.~Wang, W.~Feng, Y.~Chen, H.~Yu, M.~Huang, and P.~S. Yu, ``Visual domain
  adaptation with manifold embedded distribution alignment,'' in \emph{ACM
  international conference on Multimedia}, 2018, pp. 402--410.

\bibitem{yan2017mind}
H.~Yan, Y.~Ding, P.~Li, Q.~Wang, Y.~Xu, and W.~Zuo, ``Mind the class weight
  bias: Weighted maximum mean discrepancy for unsupervised domain adaptation,''
  in \emph{IEEE Conference on Computer Vision and Pattern Recognition}, 2017,
  pp. 2272--2281.

\bibitem{DSAN}
Y.~{Zhu}, F.~{Zhuang}, J.~{Wang}, G.~{Ke}, J.~{Chen}, J.~{Bian}, H.~{Xiong},
  and Q.~{He}, ``Deep subdomain adaptation network for image classification,''
  \emph{IEEE Transactions on Neural Networks and Learning Systems}, vol.~32,
  no.~4, pp. 1713--1722, 2021.

\bibitem{pei2018multi}
Z.~Pei, Z.~Cao, M.~Long, and J.~Wang, ``Multi-adversarial domain adaptation,''
  in \emph{AAAI Conference on Artifical Intelligence}, 2018, pp. 3934--3941.

\bibitem{GAACN}
``Generative attention adversarial classification network for unsupervised
  domain adaptation,'' \emph{Pattern Recognition}, vol. 107, p. 107440, 2020.

\bibitem{zhang2018importance}
J.~Zhang, Z.~Ding, W.~Li, and P.~Ogunbona, ``Importance weighted adversarial
  nets for partial domain adaptation,'' in \emph{IEEE Conference on Computer
  Vision and Pattern Recognition}, 2018, pp. 8156--8164.

\bibitem{cao2019learning}
Z.~Cao, K.~You, M.~Long, J.~Wang, and Q.~Yang, ``Learning to transfer examples
  for partial domain adaptation,'' in \emph{Proceedings of the IEEE Conference
  on Computer Vision and Pattern Recognition}, 2019, pp. 2985--2994.

\bibitem{chen2020selective}
Z.~Chen, C.~Chen, Z.~Cheng, B.~Jiang, K.~Fang, and X.~Jin, ``Selective transfer
  with reinforced transfer network for partial domain adaptation,'' in
  \emph{Proceedings of the IEEE/CVF Conference on Computer Vision and Pattern
  Recognition}, 2020, pp. 12\,706--12\,714.

\bibitem{Jigsaw}
S.~Bucci, A.~D'Innocente, Y.~Liao, F.~M. Carlucci, B.~Caputo, and T.~Tommasi,
  ``Self-supervised learning across domains,'' \emph{IEEE Transactions on
  Pattern Analysis and Machine Intelligence}, pp. 1--12, 2021.

\bibitem{Song2009Hilbert}
L.~Song, J.~Huang, A.~Smola, and K.~Fukumizu, ``Hilbert space embeddings of
  conditional distributions with applications to dynamical systems,'' in
  \emph{International Conference on Machine Learning}, 2009, pp. 961--968.

\bibitem{Baker1973Joint}
C.~R. Baker, ``Joint measures and cross-covariance operators,''
  \emph{Transactions of the American Mathematical Society}, vol. 186, pp.
  273--289, 1973.

\bibitem{ren2016conditional}
Y.~Ren, J.~Li, Y.~Luo, and J.~Zhu, ``Conditional generative moment-matching
  networks,'' in \emph{Advances in Neural Information Processing Systems},
  2016, pp. 2928--2936.

\bibitem{sener2016learning}
O.~Sener, H.~O. Song, A.~Saxena, and S.~Savarese, ``Learning transferrable
  representations for unsupervised domain adaptation,'' in \emph{Advances in
  Neural Information Processing Systems}, 2016, pp. 2110--2118.

\bibitem{Hu2018duplex}
L.~Hu, M.~Kan, S.~Shan, and X.~Chen, ``Duplex generative adversarial network
  for unsupervised domain adaptation,'' in \emph{IEEE Conference on Computer
  Vision and Pattern Recognition}, 2018, pp. 1498--1507.

\bibitem{lecun1998gradient}
Y.~LeCun, L.~Bottou, Y.~Bengio, P.~Haffner \emph{et~al.}, ``Gradient-based
  learning applied to document recognition,'' \emph{Proceedings of the IEEE},
  vol.~86, no.~11, pp. 2278--2324, 1998.

\bibitem{Denker1989neural}
J.~S. Denker, W.~R. Gardner, H.~P. Graf, D.~Henderson, R.~E. Howard, W.~E.
  Hubbard, L.~D. Jackel, H.~S. Baird, and I.~Guyon, ``Neural network recognizer
  for hand-written zip code digits,'' in \emph{Advances in Neural Information
  Processing Systems}, 1988, pp. 323--331.

\bibitem{Netzer2011reading}
Y.~Netzer, T.Wang, A.~Coates, A.~Bissacco, B.~Wu, and A.~Y. Ng, ``Reading
  digits in natural images with unsupervised feature learning,'' in
  \emph{Advances in Neural Information Processing Systems Workshop}, 2011.

\bibitem{peng2017visda}
X.~Peng, B.~Usman, N.~Kaushik, J.~Hoffman, D.~Wang, and K.~Saenko, ``Visda: The
  visual domain adaptation challenge,'' \emph{arXiv preprint arXiv:1710.06924},
  2017.

\bibitem{Saenko2010Adapting}
K.~Saenko, B.~Kulis, M.~Fritz, and T.~Darrell, ``Adapting visual category
  models to new domains,'' in \emph{European Conference on Computer Vision},
  2010, pp. 213--226.

\bibitem{Venkateswara2017Deep}
H.~Venkateswara, J.~Eusebio, S.~Chakraborty, and S.~Panchanathan, ``Deep
  hashing network for unsupervised domain adaptation,'' in \emph{IEEE
  Conference on Computer Vision and Pattern Recognition}, 2017, pp. 5018--5027.

\bibitem{he2016deep}
K.~He, X.~Zhang, S.~Ren, and J.~Sun, ``Deep residual learning for image
  recognition,'' in \emph{IEEE Conference on Computer Vision and Pattern
  Recognition}, 2016, pp. 770--778.

\bibitem{Kingma2014Adam}
D.~P. Kingma and J.~Ba, ``Adam: A method for stochastic optimization,''
  \emph{arVix preprint, arXiv:1412.6980}, 2014.

\bibitem{Ghifary2016Deep}
M.~Ghifary, W.~B. Kleijn, M.~Zhang, D.~Balduzzi, and L.~Wen, ``Deep
  reconstruction-classification networks for unsupervised domain adaptation,''
  in \emph{European Conference on Computer Vision}, 2016, pp. 597--613.

\bibitem{Liu2016Coupled}
M.~Y. Liu and O.~Tuzel, ``Coupled generative adversarial networks,'' in
  \emph{Advances in Neural Information Processing Systems}, 2016, pp. 469--477.

\bibitem{Sankaranarayanan2017Generate}
S.~Sankaranarayanan, Y.~Balaji, C.~D. Castillo, and R.~Chellappa, ``Generate to
  adapt: Aligning domains using generative adversarial networks,'' in
  \emph{IEEE Conference on Computer Vision and Pattern Recognition}, 2018, pp.
  8503--8512.

\bibitem{YANG2021107638}
``Learning adaptive geometry for unsupervised domain adaptation,''
  \emph{Pattern Recognition}, vol. 110, p. 107638, 2021.

\bibitem{long2017deep}
M.~Long, H.~Zhu, J.~Wang, and M.~I. Jordan, ``Deep transfer learning with joint
  adaptation networks,'' in \emph{International Conference on Machine
  Learning}, 2017, pp. 2208--2217.

\bibitem{Maaten2008Visualizing}
L.~V.~D. Maaten and G.~Hinton, ``Visualizing data using t-{SNE},''
  \emph{Journal of Machine Learning Research}, vol.~9, pp. 2579--2605, 2008.

\end{thebibliography}

\clearpage


\end{document}